\documentclass{article}

\usepackage{arxiv}

\usepackage{lineno,hyperref}
\modulolinenumbers[5]
\usepackage{xcolor}
\usepackage{graphicx}
\usepackage{subfig}
\usepackage{amsfonts}

\usepackage{booktabs}
\usepackage{amssymb}
\usepackage{algorithm}
\usepackage{algpseudocode}
\usepackage{hyperref} 
\usepackage[normalem]{ulem}  

\title{Incremental Learning from Low-labelled Stream Data in Open-Set Video Face Recognition}

\author{\and\href{https://orcid.org/0000-0002-7720-1607}{\includegraphics[scale=0.06]{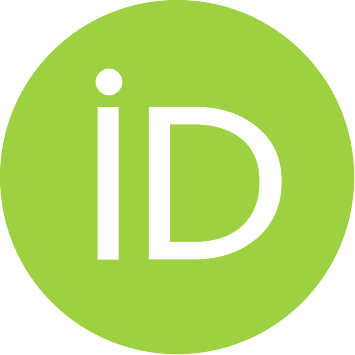}\hspace{1mm}Eric Lopez-Lopez}\\
	CITIC\\
	Universidade da Coruña\\
	\texttt{eric.lopez@udc.es} \\
	\and
	\href{https://orcid.org/0000-0003-3672-1726}{\includegraphics[scale=0.06]{orcid.pdf}\hspace{1mm}Carlos V. Regueiro} \\
	CITIC\\
	Universidade da Coruña\\
	\texttt{carlos.vazquez.regueiro@udc.es} \\
	\and
	\href{https://orcid.org/0000-0002-3997-5150}{\includegraphics[scale=0.06]{orcid.pdf}\hspace{1mm}Xose M. Pardo} \\
	CiTIUS\\
	Universidade de Santiago de Compostela\\
	\texttt{xose.pardo@usc.es} \\
}

\hypersetup{
pdftitle={Incremental Learning from Low-labelled Stream Data},
pdfsubject={cs.CV},
pdfauthor={Eric Lopez},
pdfkeywords={First keyword, Second keyword, More},
}

\begin{document}
\maketitle

\begin{abstract}
	Deep Learning approaches have brought solutions, with impressive performance, to general classification problems where wealthy of annotated data are provided for training. In contrast, less progress has been made in continual learning of a set of non-stationary classes, mainly when applied to unsupervised problems with streaming data.

    Here, we propose a novel incremental learning approach which combines a deep features encoder with an Open-Set Dynamic Ensembles of SVM, to tackle the problem of identifying individuals of interest (IoI) from streaming face data. From a simple weak classifier trained on a few video-frames, our method can use unsupervised operational data to enhance recognition. Our approach adapts to new patterns avoiding catastrophic forgetting and partially heals itself from miss-adaptation. Besides, to better comply with real world conditions, the system was designed to operate in an open-set setting. Results show a benefit of up to 15\% F1-score increase respect to non-adaptive state-of-the-art methods.
\end{abstract}


\section{Introduction}

Deep Learning approaches have brought solutions, with impressive performance,  to general classification problems where wealthy of annotated data are provided for training. Given the fact that in real-world applications specific data are many times scarce, very costly to label, non-stationary,  or streaming, new and classical learning strategies have been incorporated to the realm of Deep Learning, to deal with these challenges \cite{kemker2018measuring}. Thus, topics as transfer learning \cite{wang2018deep}, 
 reinforcement learning \cite{Ren2018DeepRL}, 
or incremental learning \cite{he2020incremental,sahoo2018online,tao2020fewshot,perezrua2020incremental}, both supervised and unsupervised, have gained new momentum.

Incremental learning is the ability of a classifier to evolve by continuously integrating information from new instances and/or new classes, and without resort to full retraining \cite{kemker2018measuring}. Currently, incremental and online machine learning are receiving more and more attention especially in the context of learning from real-time data streams \cite{he2020incremental,sahoo2018online}. In particular, rehearsal-free continual learning techniques have also demonstrated their abilities to extend the class-set of a classifier considering only labels from the new classes,  while avoiding the problem of catastrophic forgetting \cite{tao2020fewshot,perezrua2020incremental}. That is of special interest when computational capacities do not allow full retraining, or confidentiality issues impede new access to old samples during the process of extending the class set. In contrast,  less progress has been made in continual learning of a set of non-stationary classes, mainly when applied to tasks involving unsupervised streaming data.

A paradigmatic example of the application of incremental learning, dealing with unsupervised, non-stationary and streaming data is the case of video-to-video face recognition (V2V-FR) in video surveillance \cite{huang2015cox}. 
Usually, video-frame are captured with a broad range of individual pose,  camera position, resolution, and illumination, which often excess the diversity available  in datasets used to train deep networks (generally focused on web extracted images) \cite{guo2016mscleb1m}
. Transfer learning to specific task domains in V2V-FR has proven to be challenging even for Deep Learning encoders \cite{gunther2016face,lopezlopez2019dataset} since image quality factors are still decisive for performance \cite{guo2019survey}
.  Besides, since the data are received continuously in a stream fashion, individual appearance could change when switching between different cameras, which could also operate in changing conditions over time \cite{ditzler2015learning,disabato2019learning}. While, in theory, all of these issues can be solved with further labelling, the task of having addressed every possible variation in a training dataset is, in practical terms, infeasible \cite{he2020incremental}. Then, a more efficient and scalable approach is needed \cite{maltoni2019continous,tao2020fewshot,perezrua2020incremental}. In this regard, what truly represents the context of application and the changes that appear over time is the actual data \textbf{incrementally} extracted during the operation of the system, and so \textbf{without labels}.

\begin{figure}[t]
    \centering
    \includegraphics[width=\textwidth]{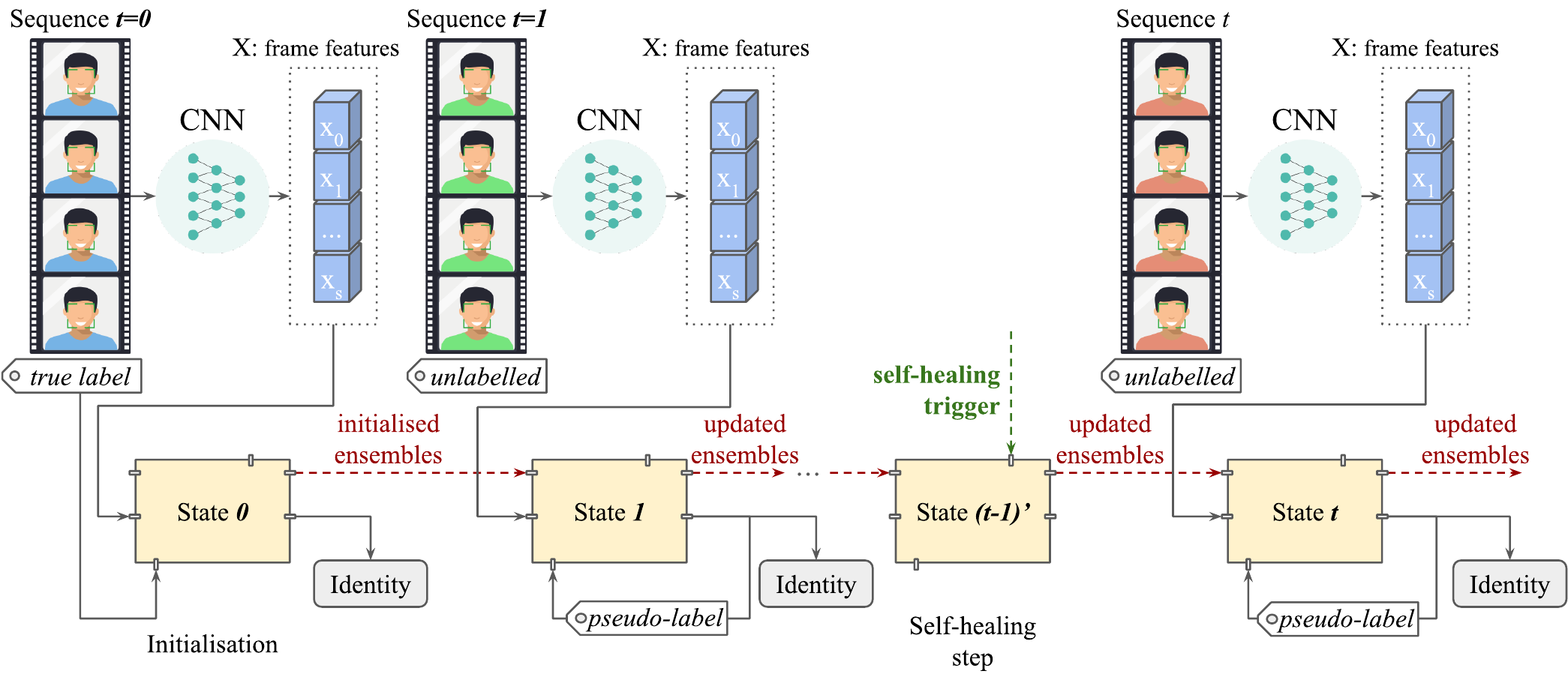}
    \caption{Open-Set Dynamic Ensembles of SVM (OSDe-SVM) is able to incorporate new knowledge and correcting wrong updates by adding and removing classifiers in an unsupervised way. The system is designed to work under open-set recognition conditions.}
    \label{fig:illustrate}
\end{figure}

Another characteristic of real-world applications of V2V-FR relates to their intrinsic open-set nature \cite{scheirer2013toward,gunther2017towards}
. Take for example a practical case of an airport video-surveillance aimed to track some \emph{individuals of interest (IoI)} who have not been collaboratively enrolled in the system, e.g.\ those exhibiting suspicious behaviours, among a larger number of \emph{unknown} non-target identities, that should be identified as \emph{unknown} non-target identities. 

In this paper, we propose a novel incremental learning system, the Open-Set Dynamic Ensembles of SVM \ref{fig:illustrate}. Using deep feature encoder as a basis, the system is capable of using operational data to enhance and improve the recognition of target identities in an unsupervised way. Additionally, guided by the real-world necessities, the system is designed to operate in a completely open set setting. Rooted on the power of a Deep features encoder, trained for the general face recognition problem, an incremental learning module, fed with stream data, simultaneously predict and update classifiers, while dealing with catastrophic forgetting issues. The incremental module is based on dynamic ensembles of SVM classifiers, which from a single SVM built from a few labelled video-frames directly taken from the footage, can acquire and adapt to additional information by adding/removing classifiers to/from ensembles. It follows the self-training strategy \cite{yarowsky1995unsupervised} in which predictions also play the role of pseudo-labels, which are used to update and improve the classifiers. Based on the modular nature of ensembles, adaptations consisting of either adding or removing classifiers. Contributions of this paper can be summarised as:

\begin{itemize}
    \item An approach to unsupervised incremental learning designed to operate online with stream data. During its operation, predictions also play the role of pseudo-labels.
    \item A strategy to deal with both catastrophic forgetting issues and the effect of mistaken pseudo-labels.
    \item An approach to instance-incremental learning in the open-set, which could be extended to cope with the class-incremental problem.

    \item A method for person re-identification based on face, which is not directly based on a reservoir of face images.
    
\end{itemize}

The rest of the paper is organised as follows. First, in Sec.~\ref{sec:rel_work} we perform an extensive study of the existent literature related to the problem. After that, we move to present the proposed approach in Sec.~\ref{sec:desvm} and a set of experiments to study its behaviour, in Sec.~\ref{sec:experiments} and~\ref{sec:results}. Finally, in Sec.~\ref{sec:conclusion}, we reflect on the conclusions we can extract from the work.

\section{Related Work}
\label{sec:rel_work}

\textbf{Open-set Recognition.} 
In open set recognition, training is performed on a dataset with samples of some known classes, while samples of both known and unknown classes are presented for testing. Therefore, classifiers should appropriately deal with all of them. 
Within this approach, closer to real-world applications, decision boundaries not only separate instances of different known classes, but they separate the known from the unknown as well \cite{fayin2005open,scheirer2013toward}
. A recent survey \cite{geng2020recent} distinguishes between discriminative and generative approaches to open set recognition. Discriminative classifiers are trained to discriminate between the known classes, and then, given the most likely class label, to decide whether a test sample was in fact drawn from the distribution of known class samples or not \cite{gunther2017towards}. Meanwhile, generative methods try to provide explicit probability estimation over unknown categories, most of them based on deep networks \cite{zongyuan2017generative,perera2020}. Plenty of methods in both sets of approaches, leverage Extreme Value Theory (EVT) to tackle the unknown \cite{coles2001evt}. 
EVT is a branch of statistics aimed to assess the probability of observing an event more extreme than any previously observed, and has been widely used for outlier detection 
in open-set recognition \cite{rudd2018evm}. 

In face recognition, the most realistic scenario corresponds to an open-set setting (e.g. criminal watch-lists, restricted areas access control, smart-homes, etc.) \cite{gunther2017towards}. In this domain, apart from EVT based methods, solutions based on siamese networks have been proposed to address the open-set as they are metric learning methods, and their similarity scores can be thresholded to perform  recognition \cite{salomon2020openset}. Although they do not fit the data stream context, they could be used as a baseline for comparison purposes \cite{schroff2015facenet}.

\textbf{Incremental Learning.} The main goal of incremental (a.k.a\ lifelong, continuous or continual) learning is to learn from data as they are provided by real-world dynamic sources, usually at a low pace, including noisy samples and, in general, exhibiting non-stationarity. As data distributions change with time, computational systems have to deal with the \emph{stability-plasticity dilemma}, trying to avoid that new knowledge erases old one (\emph{catastrophic forgetting}), while detecting and updating to concept or data drifts \cite{sahoo2018online}.  

In the context of deep approaches, continual learning has been focused on learning new tasks/classes, more than on enhancing the performance of classifiers (fixed number of classes) as new instances arrive \cite{maltoni2019continous,parisi2019}. Among common strategies are the exploitation of, at least, partial rehearsal (looping over old data) \cite{kemker2018measuring,hayes2020lifelong}
, dynamic changes in architectures (retraining after pruning/increasing the number of neurons, filters or layers), and regularisation (updating weights in order not to forget previous knowledge) \cite{kirkpatrick2017}. Among the last are usually also included a wide range of knowledge distillation methods, in which a teacher network transfer knowledge to a student network \cite{li2018learning, castro2018endtoend, zhang2020classincremental}. However, the drawback of distillation is that it generally needs to retain big past memories \cite{belouadah2020scail, tao2020fewshot}. Notwithstanding the progress made in supervised incremental learning in the recent years, there is still a substantial gap between the performance of batch offline learners on stationary data and the performance of the incremental learners that deal with non-stationary data \cite{hayes2020lifelong, kemker2018measuring}. 

Most of the work carried out to date regarding incremental learning is focused on batches. So, they need to wait for a batch of data to accumulate before a new adaptation can take place. Only a bunch of approaches were really designed to tackle the problem of incremental learning from streaming data, which is considered a more challenging task \cite{vandeVen2020brain}
. One of its critical difficulties is the infeasibility of complete manual labelling of streaming data in real-world applications. A more realistic approach should only assume that a few instances in data streams are labelled \cite{salah2020}. 

Most of the semi-supervised methods leverage unlabelled examples by making some assumptions, using label propagation or generating pseudo-labels during the learning process \cite{li2019}. Some approaches are based on keeping a set of dynamic clusters to summarise class distributions and model their evolution overtime \cite{salah2020}. Others use a few labelled data to initialise a set of models, which are afterwards sequentially updated based on pseudo labelled data \cite{delatorre2015partially,pisani2019adaptive,orru2020novel}. In the specific case of video recognition, weak labels can be provided by the temporal tracking \cite{franco2010incremental,pernici2017unsupervised}, but also co-training or predictions of the own classifiers can provide pseudo-labels.

Ensemble methods have been acknowledged as powerful tools to overcome \emph{catastrophic forgetting} \cite{polikar2001learn,coop2013ensemble,kemker2018measuring}, when dealing with data streams \cite{krawczyk2017ensemble,delatorre2015partially}. Moreover, ensemble algorithms can be integrated with drift detection algorithms and incorporate dynamic updates, such as selective removal or addition of classifiers \cite{gomes2017}. In the semi-supervised scenario, it must be taken into account that any kind of weak labelling or pseudo labelling is prone to error. So, dynamic updates can be also useful to healing form the effect of mislabelling. Unlike other incremental learning approaches (either classic \cite{kivinen2002online,liang2006fast} or DL-based \cite{he2020incremental}), ensembles provide simple way to isolate updates and, consequently, make changes reversible. And not only that, since decisions are based on majorities, ensembles are robust to outliers. 

In \cite{dvornik2019diversity} ensembles of deep networks have been proposed to encourage networks to cooperate and take advantage of their prediction diversity, in the context of few-shot classification. Besides, to deal with tasks where training data are inadequate, the training of a collection of incrementally fine-tuned CNN models and their combination using an ensemble, was presented \cite{zhang2020}. In \cite{guo2020}, the authors propose an ensemble learning framework based on multiple CNN classifiers. The CNN acts as a feature extractor for the posterior use of different ensemble frameworks to classify its content. Recently, already in the context of incremental learning, an approach based on ensembles, which is close to ours, was proposed for tackling the problem of mechanical fault diagnosis \cite{wang2020}.

Although there are propositions of end-to-end deep learning approaches for incremental semi-supervised learning \cite{li2019, orru2020novel}, their inherent characteristics make them yet unsuitable to operate online with streaming data. Therefore, for this specific context, we propose to combine the good characteristics of a deep feature encoder, which transfers knowledge from the source domain, with an ensemble method able to provide adaptation to the target domain.

\section{Proposed Method: Deep Embeddings + Open-Set Dynamic Ensembles of SVM (OSDe-SVM)}
\label{sec:desvm}

\begin{figure}
    \centering
    \includegraphics[width = \textwidth]{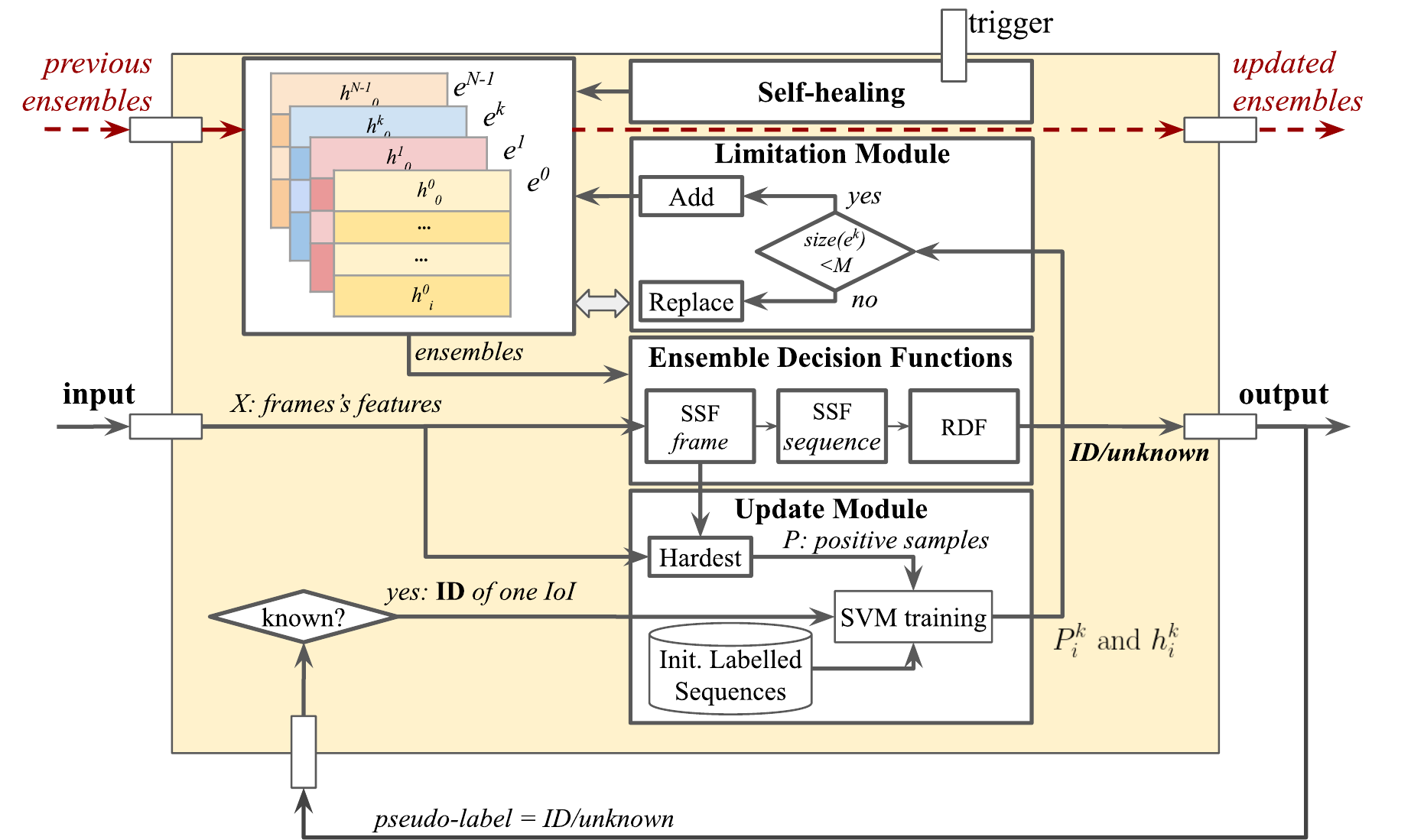}
    \caption{Pipeline of OSDe-SVM.}

    \label{fig:pipeline}
\end{figure}

In this work we present the \emph{Open-Set Dynamic Ensemble of SVM (OSDe-SVM)} for the problem of V2V-FR in the open-set context (Fig.~\ref{fig:illustrate}). This method takes advantage of transfer learning from large labelled datasets, to get discriminant feature embeddings that feed an instance-incremental learning module. The complete method is able to achieve online adaptation to the task domain from unlabelled streaming data. To do so,  OSDe-SVM relies on the self-training strategy and the modular nature of ensembles to add and remove classifiers in a totally autonomous way.

OSDe-SVM uses features taken after the last pair of convolutional and batch normalisation layers of the ResNet100-ArcFace (RN100-AF) network trained on MS1MV2 dataset \cite{deng2019arcface}. This is one of the top-performing CNN in the face recognition state-of-the-art. ArcFace is a loss-function specifically designed to enhance the discriminative power of face recognition models, being the deepest networks, as ResNet-100, the ones that take the most advantage of it \cite{deng2019arcface}. The encoding transforms a 112x112 face crops into a 512-D feature embedding. 

The general structure of OSDe-SVM is depicted in Fig.~\ref{fig:pipeline}. Each individual of interest (IoI), $k$, has an associated ensemble, $e^k$, composed by a set of SVM classifiers, $h_i^k$. This ensemble is updated whenever the system is queried. The update mechanism consists of adding classifiers basing on the Ensembles Decision Functions, Sec.~\ref{sec:decisions}, following the self-training paradigm (Sec.~\ref{sec:incremental}). Besides, OSDe-SVM can remove classifiers when the maximum number of classifiers is reached (Limitation Module, Sec.~\ref{sec:limit}) or when a possible mistake is detected (Self-healing, Sec.~\ref{sec:selfhealing}). Each SVM classifier is trained with a few amount of positive samples (face crops of the first 5 frames containing each IoI) against a pool of samples from the target domain. 

\subsection{Ensemble Decision Functions}
\label{sec:decisions}

OSDe-SVM builds, and keeps updated, ensembles aimed at the re-identification of each IoI within the area of a camera network  (Fig.~\ref{fig:pipeline}). Ensemble's decisions are made in a two-step process. First, the \emph{Sequence Scoring Function} assigns a certain score to the query sequence. And second, the \emph{Recognition Decision Function} uses these scores to assign an identity label (either as one of the IoI or as an unknown).

\subsubsection{ Sequence Scoring Function}

When making decisions, it is convenient that each ensemble gives a unique score to each incoming sequence. Nevertheless, both (sequences and ensembles) are composed elements. Being $n_F$ the number of sequence's frames and $M^k$ the number classifiers of ensemble $k$, we would have a total of  $n_F\times M^k$ different responses. We call the \emph{Sequence Scoring Function (SSF)} to the process of combining all of these different responses into a unique score. This process consists of two levels:

\begin{itemize}
    \item At \emph{frame level}, we combine the responses of the ensemble's classifiers to give a unique score to each frame. The function used here is the median of the individual ensemble's SVM scores. In practice, this corresponds to a majority voting.
    
    \item At \emph{sequence level}, we take advantage of the temporal coherence assumption to assign a unique identity to the whole input sequence. This assumption allows us to combine all the frame's scores into a unique one. The function used here is the median.
\end{itemize}

\subsubsection{ Recognition Decision Function based on Extreme Value Theory }
\label{sec:rdf_evt}

Once every ensemble delivers its prediction score about an input query,
the next step is to combine all the predictions to decide the underlying identity. The identity assignment based on the best score is the usual procedure in a closed-set scenario  \cite{deng2019arcface, liu2019adaptive}. That is because input sequences always belong to a known IoI. In an open-set scenario, assigning identities becomes trickier because \emph{non-match responses}, corresponding to unknown identities, are also expected. To tackle these scenarios, OSDe-SVM was endowed with a \emph{Recognition Decision Function (RDF)} based on EVT, which also allows dealing with the uncalibrated outputs provided by SVM's.

\begin{figure}[t]
    \centering
    \includegraphics[trim=30 0 10 0,width=0.325\textwidth]{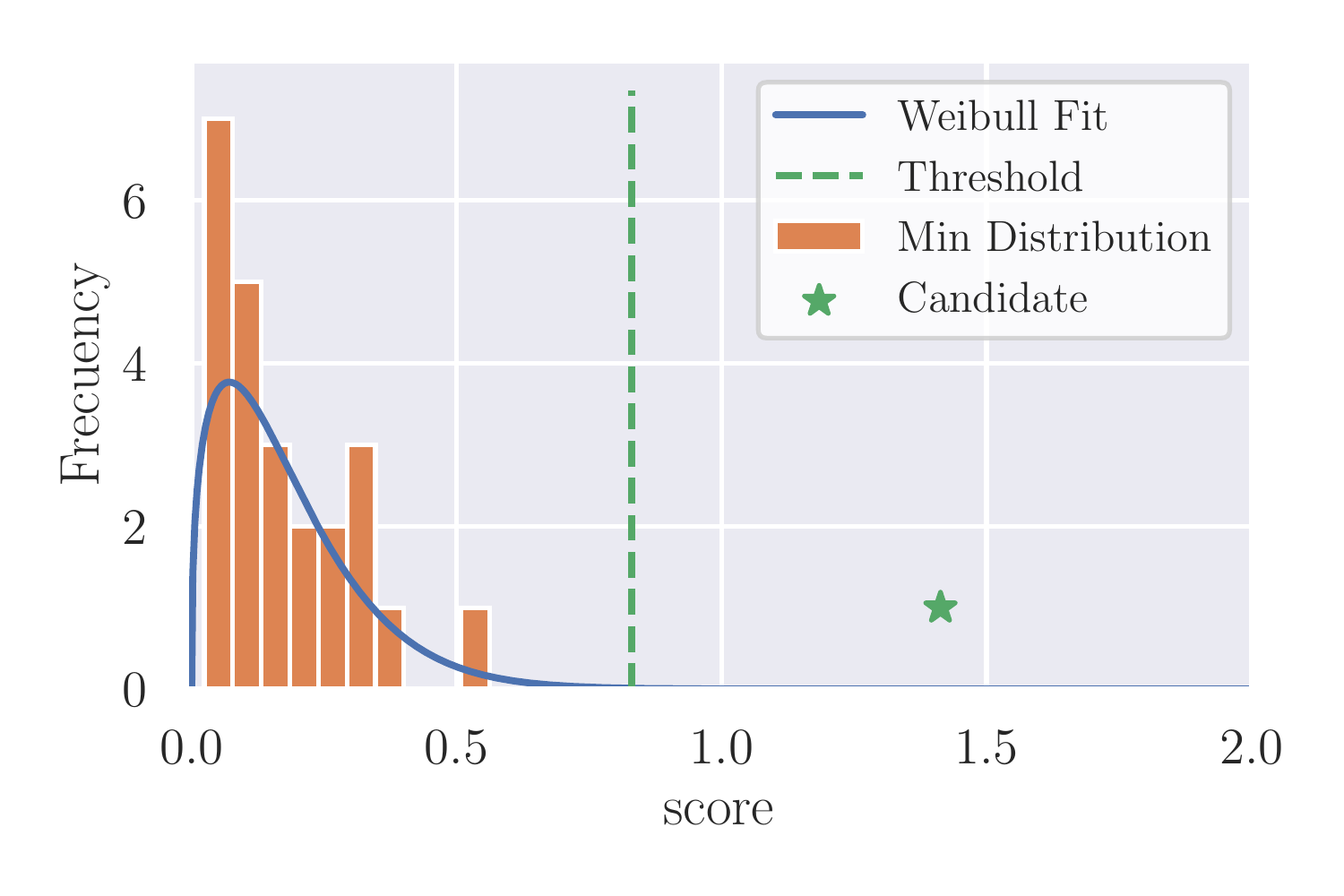}
    \includegraphics[trim=30 0 10 0,width=0.325\textwidth]{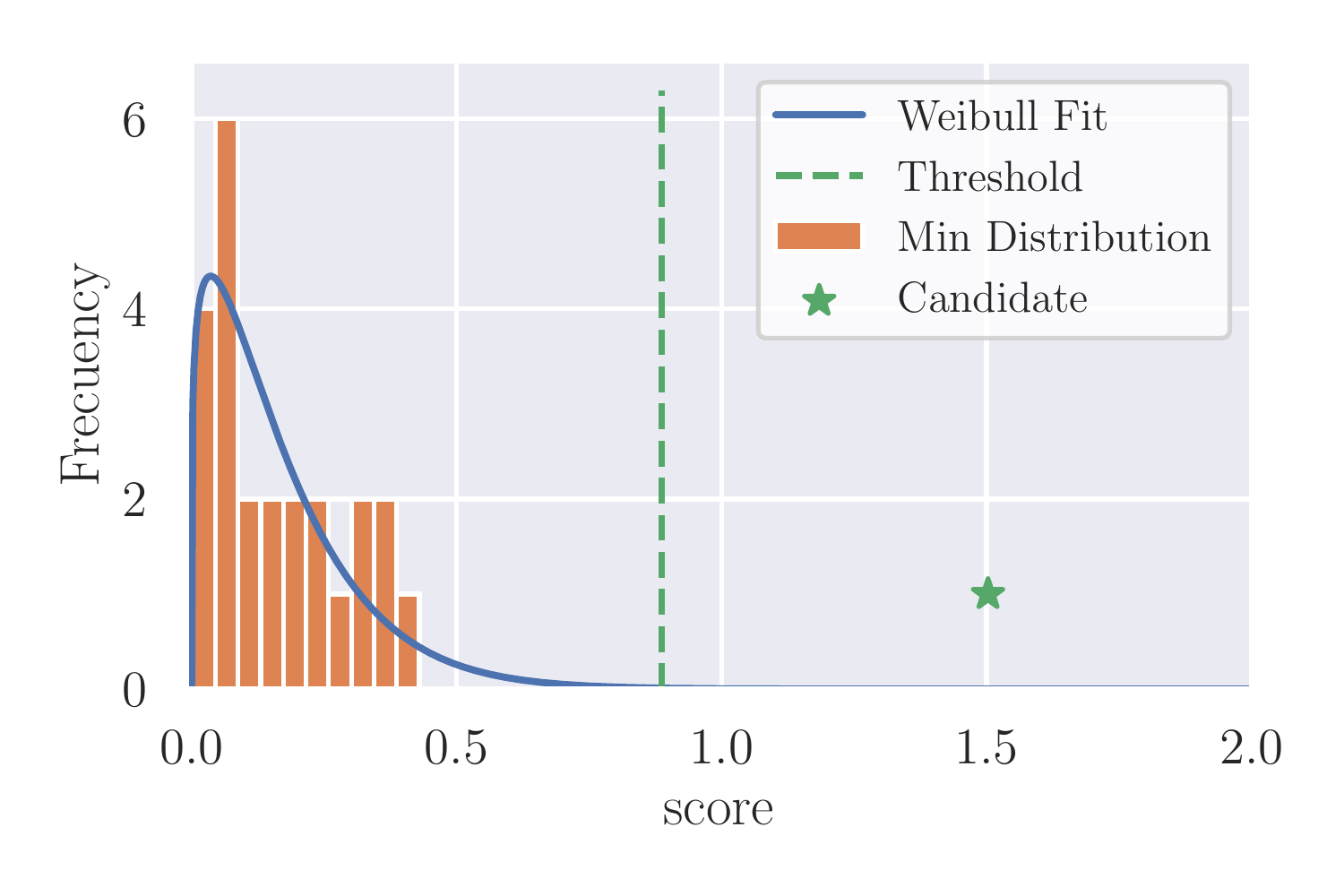}
    \includegraphics[trim=30 0 10 0,width=0.325\textwidth]{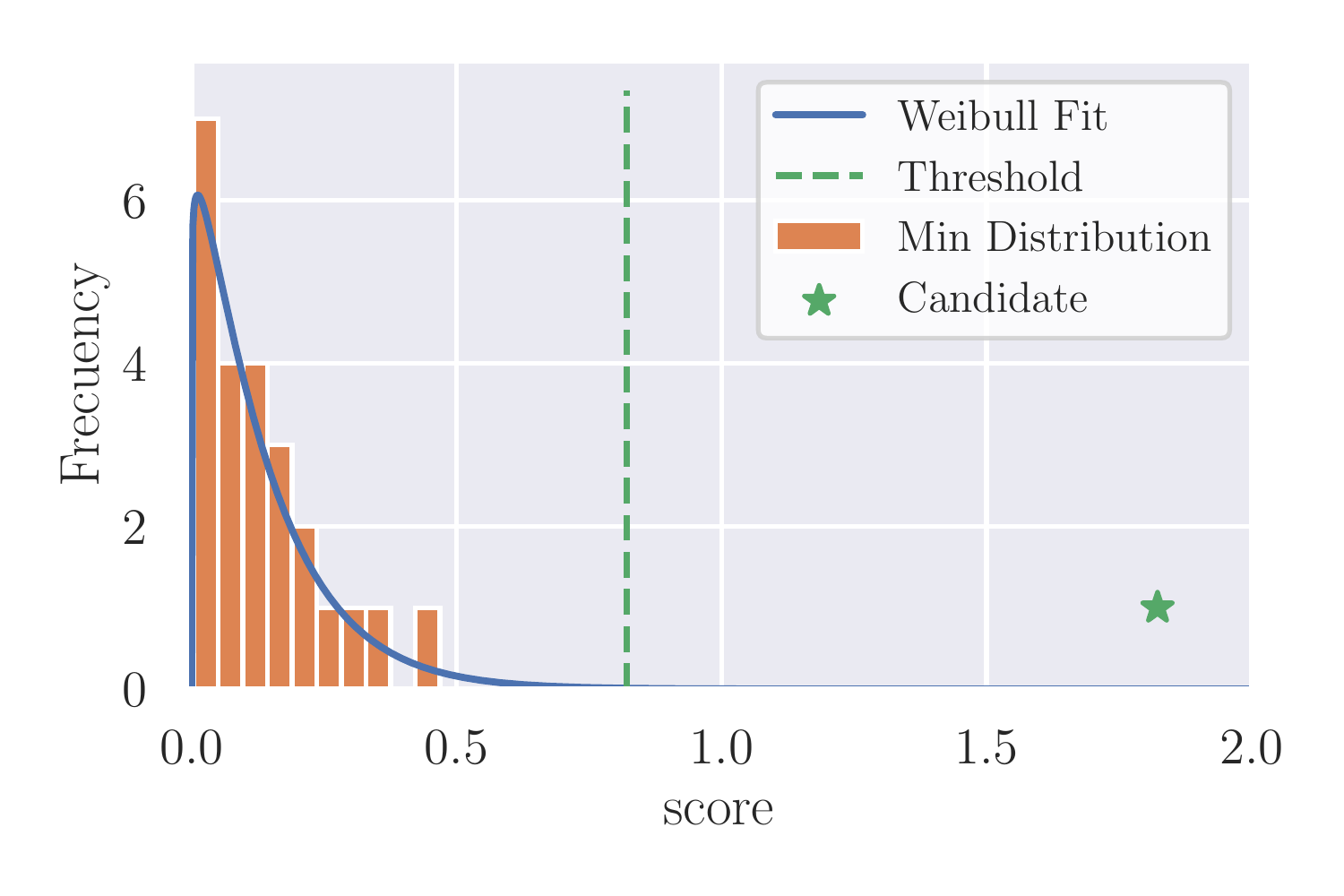}
    \includegraphics[trim=30 0 10 0,width=0.325\textwidth]{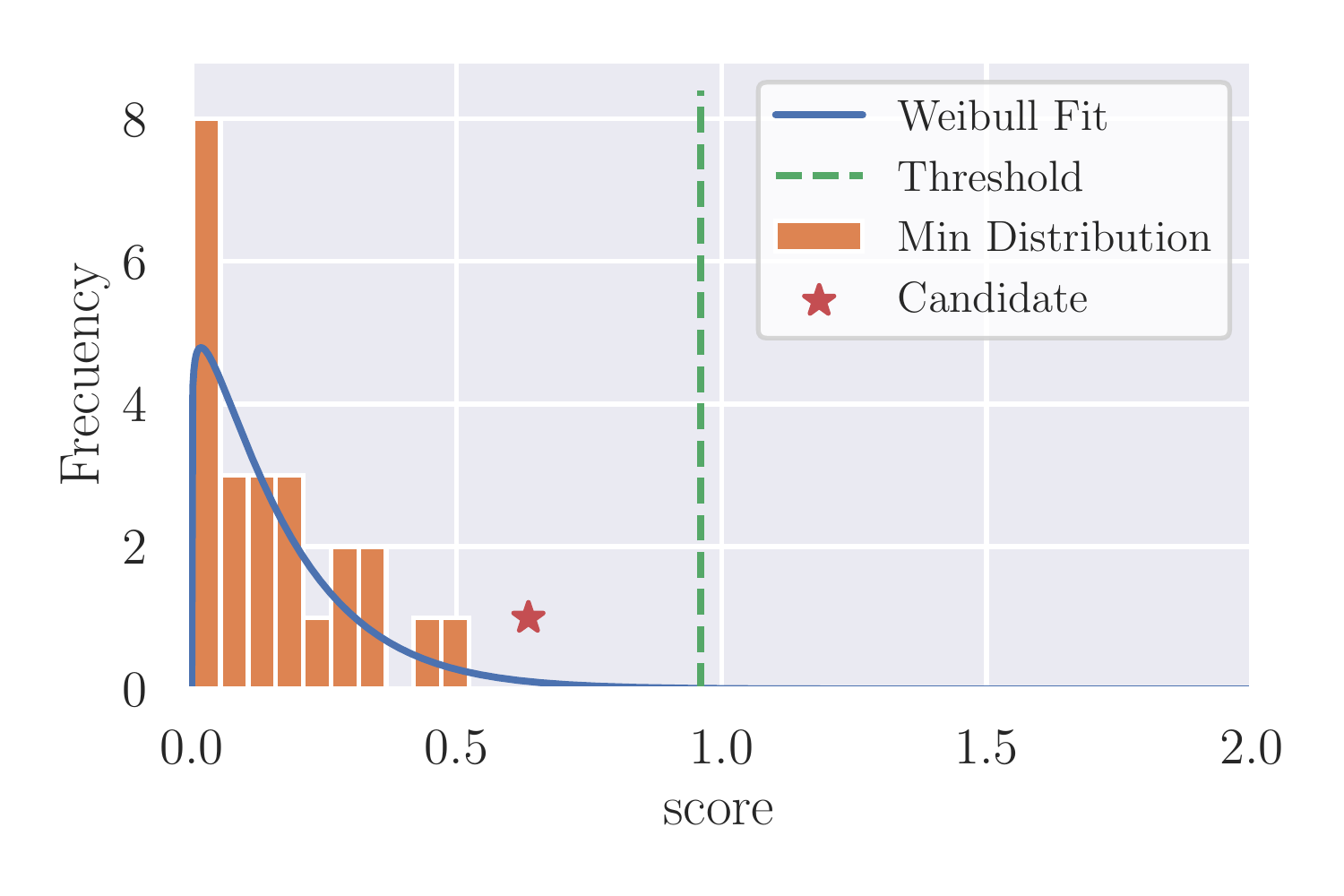}
    \includegraphics[trim=30 0 10 0,width=0.325\textwidth]{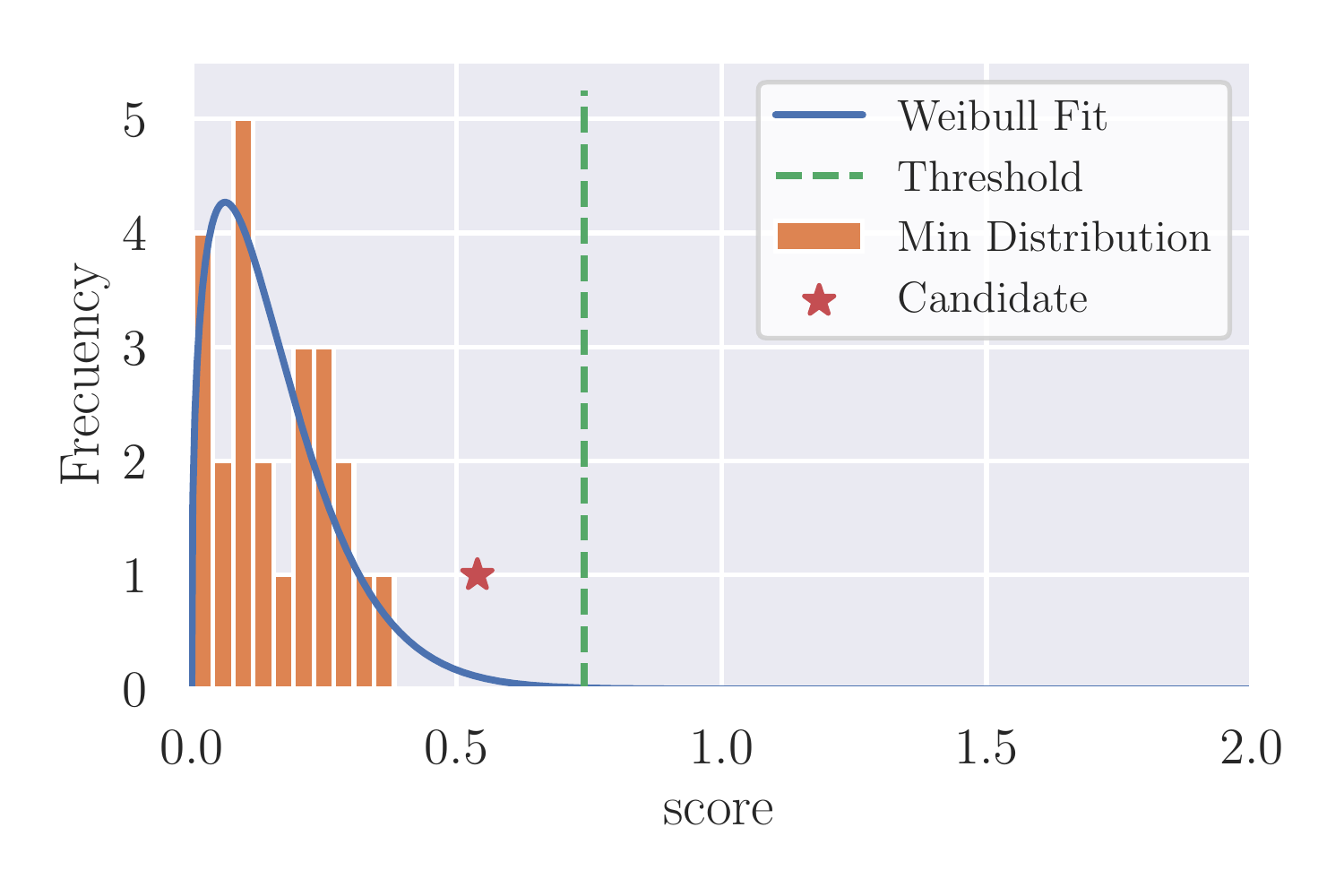}
    \includegraphics[trim=30 0 10 0,width=0.325\textwidth]{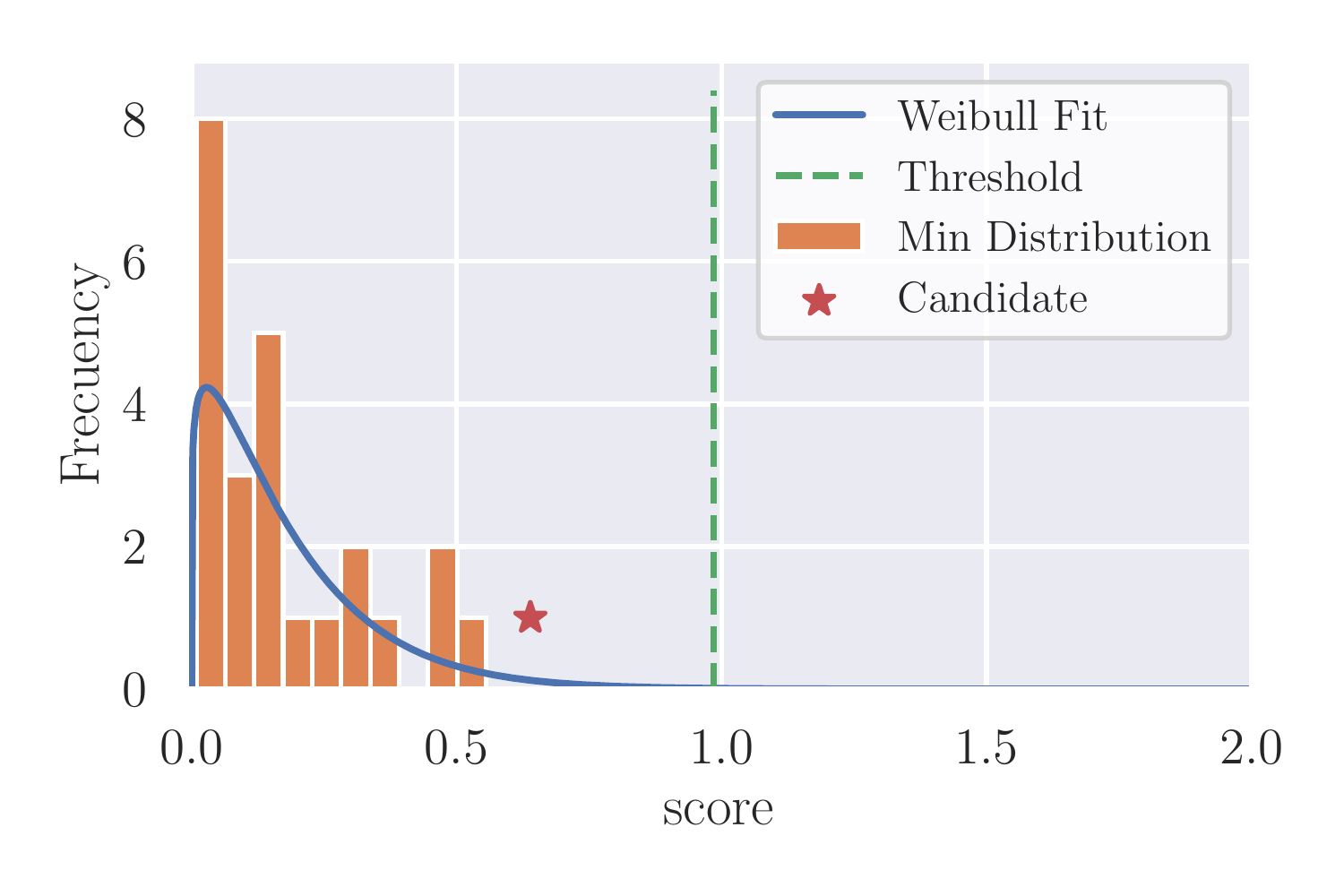}
    \caption{Examples of Extreme Values distributions and application of RDF. A Weibull function is fitted to the min distribution to see whether the candidate belongs or not to this distribution. First row illustrates examples of \emph{known} identities and second row does the same with the \emph{unknwon} ones. 
    }
    \label{fig:evt}
\end{figure}

When applying EVT, we follow an approach similar to \cite{scheirer2010robust}. As any input sequence belongs to a unique identity, the ensembles associated with other identities should deliver \emph{non-match} outputs. According to the Fisher-Tippet-Gnedenko Theorem of EVT \cite{coles2001evt}, the distribution of these \emph{non-match} scores is modelled by some particular functions.

In this case,  for left bounded positive samples, the distribution of minima, $G(z)$, is given by the Reversed Weibull distribution. We can perform a simple transformation to the ensemble's scores (given by the SSF) to satisfy these conditions and be able to fit a Weibull distribution to the tail of the distribution, as is depicted in Alg.~\ref{algo:evt}. Then, to discriminate between \emph{unknown} and \emph{known} identities,  the best ensemble response (the best score) can be checked whether it comes from this \emph{non-match} distribution (Fig.~\ref{fig:evt}) or not.

More importantly, Alg.~\ref{algo:evt} also provides a way of distinguishing the known from the unknown by thresholding the Weibull distribution ($T_W$), instead of the actual scores. Since the fitted function is different depending on the input sequence, we are implicitly personalising the threshold to each input sequence, as is depicted in Fig.~\ref{fig:evt}.  

\begin{algorithm}[t]
\small
  \begin{algorithmic}[1]
\State $S$ is the input sequence, $T_W$ is the threshold in the Weibull function
\State $E= \left\{e^0, e^1, \dots, e^{N-1}\right\}$ set of ensembles associated to \emph{known} identities
\State $ R =  \left \{ \varnothing \right \}$ set of scores given by each ensemble to a candidate
\For {$e^i$ in $E$} 
    \State $R \gets SSF(e^i,S)$
\EndFor

\State $c = \min(R)$; $m = median(R\setminus c)$
\State $V = \left\{\left\|x-m\right\| \mid x\in (R\setminus c) \wedge  (x<m)\right\}$
\State Fit V to a Weibull function, $W$ 
\If {$W(\left\|c-m\right\|) < T_W$}
    \State ID $= arg(c)  ~~$
\Else
    \State ID $=$\emph{unknown}
\EndIf

\caption{
Recognition Decision Function (RDF) based on EVT.
\label{algo:evt}}
\end{algorithmic}
\end{algorithm}

\subsection{Update Module: Incremental learning based on Self-Updating }
\label{sec:incremental}

OSDe-SVM was conceived to operate in the context of a shortage of labelled data. Only the first classifier of each ensemble is trained with a very short labelled sequence extracted from the input. The first five frames have proven to be the bare minimum for our method. From that point on, incremental learning is exclusively based on pseudo-labels (Fig.~\ref{fig:illustrate}).

After an ensemble is initialised, our method decides whether a new classifier must be added up to enhance future performance, each time a sample of the same identity is identified.  OSDe-SVM follows a self-updating strategy based on pseudo-labels provided by EDF to input sequences. Whenever an identity, $k$, is identified in an input sequence, a new SVM is created using as (pseudo-labelled) positive samples, $P_j^k$, the  5 \textbf{hardest} frames of the sequence, namely those which got the lowest scores returned by the SSF (see Fig.~\ref{fig:pipeline}). This way, diversity within each ensemble is encouraged.

\subsection{Limitation Module} 
\label{sec:limit}

In a self-updating context where each ensemble is initialised with only one classifier trained with a few labelled frames,  further updates can only occur when close samples of the same identity query the system. If they are almost identical, there is nothing to be learnt. However, if they are very different, there is a danger of not being identified. So, the model can only learn from samples in the borderline, i.e.\ samples that can still be recognised by the ensemble of the corresponding identity, but which also include some level of novelty in their features. However, ensembles' size should not grow indefinitely whenever the \emph{EDF} recognised their target identities in input sequences. As ensembles' performance relies on diversity, 
we have chosen a solution inspired in \cite{nan2012diversity}, to decide which classifiers are to be removed once maximum size is reached.

Classifies are compared against each other to obtain a measurement of their relative relevance, the  
\emph{diversity score} $D(\cdot)$. Given an ensemble, $e^k$, composed by $M_k$ SVM classifiers, $\left\{h_0^k, h_1^k,...,h_{M_k-1}^k\right\}$, $D(h_i^k)$, is computed from the binary response of each of the classifiers of the ensemble over a certain set of video frame features $\left\{x_0,x_1,...,x_{Q-1}\right\}$:

\begin{equation}
    D(h_i^k) = \sum_{j=0; j\neq i}^{N-1} d\left(h_i^k,h_j^k\right) 
\end{equation}

\begin{equation}
    d\left(h_i^k,h_j^k\right) = -\frac{1}{Q}\sum_{q=0}^{Q-1} sgn\left(h_i^k(x_q)\right) \cdot sgn(h_j^k(x_q)),
\end{equation}
where $h_i^k\left(x_q\right)$ is the response of the SVM classifier $h_i^k$ to the frame feature $x_q$, and $sgn(\cdot)$ is the sign function.

Whenever an ensemble $e^k$ reaches the maximum size, the classifier $h^k_*$ with the lowest diversity will be removed.

\subsection{Self-healing: Correcting Wrong Updates}
\label{sec:selfhealing}

\begin{figure}
    \centering
    \includegraphics[width = \textwidth]{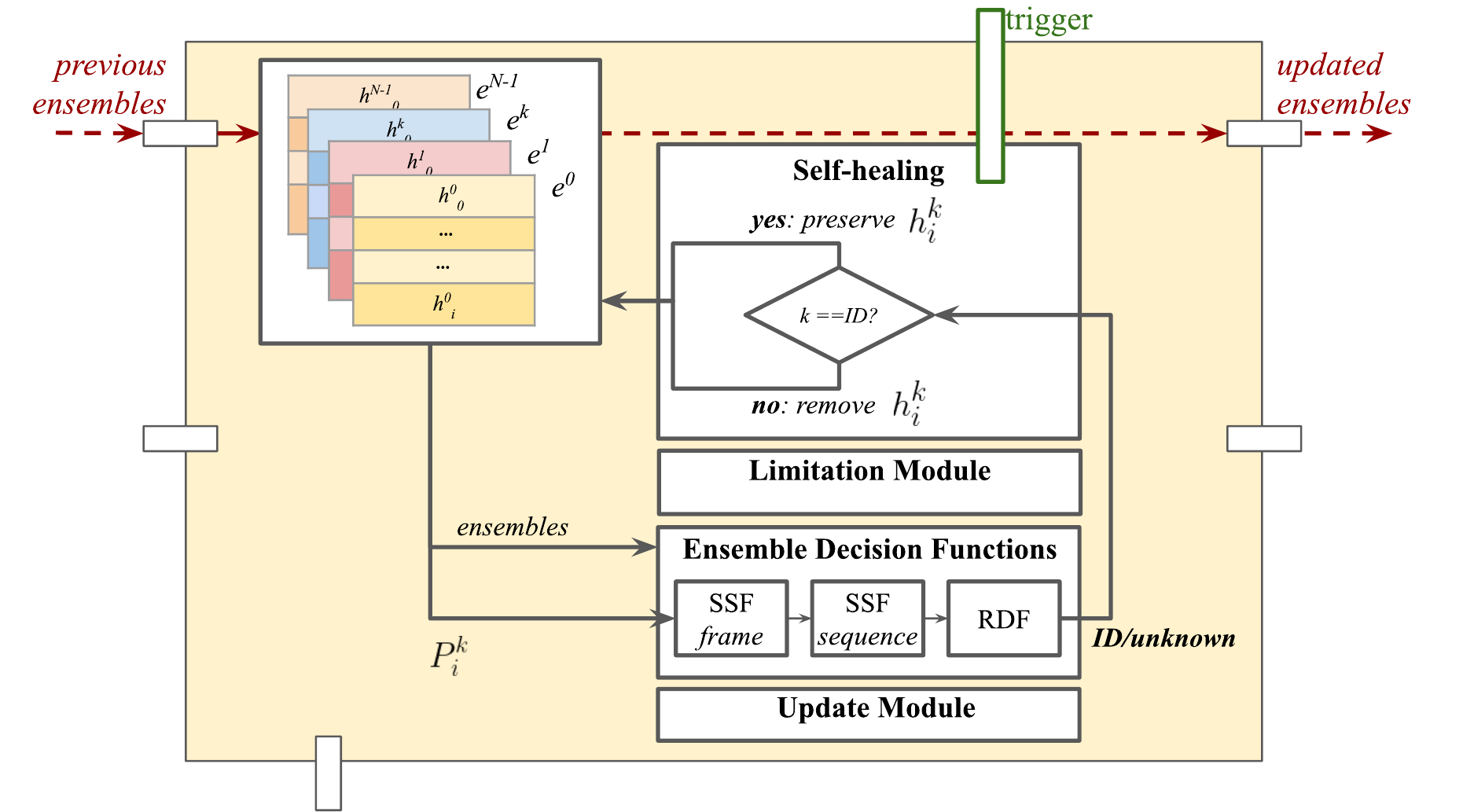}
    \caption{Pipeline of OSDe-SVM when self-healing is performed.}
    \label{fig:pipeline_selfhealing}
\end{figure}

Since the whole adaptation process performs \textbf{without supervision}, wrong updates, provoked by errors in pseudo-labelling, should be expected. This behaviour may affect re-identification performance, mainly in the long term. The \emph{self-healing} procedure is designed to mitigate this problem.

Self-healing relies on the fact that the ensembles build its decisions based in majorities. Therefore, if an ensemble reaches a relatively high accuracy in the first classifications, it should be difficult for wrong classifiers to take over very soon. This fact opens the possibility of detecting wrong updates before it becomes irreversible. We expect that, with a limited amount of wrong updates, ensembles are still able to recognise their target identity. Consequently, the future detection of the target identity can build a stronger majority capable of detecting the previous wrong update.  

To implement these ideas, along with each SVM classifier, $h_i^k$, we store the positive samples used to create it, $P_i^k$, which, in practice, can be considered a sequence. Therefore, we can pass every set (for all $k$ and $i$) again through the \emph{EDF} for a re-evaluation. If the system assigns the same identity as before, the classifier is maintained. Otherwise, the classifier is removed (Figs.~\ref{fig:pipeline_selfhealing}).  
The self-healing module triggers after a certain period which is adjustable (see Fig.~\ref{fig:illustrate}).

\section{Experimental Preliminars}
\label{sec:experiments}

\subsection{Database Selection}
Frames' quality (especially in terms of resolution), which can vary substantially depending on the context, has a direct impact on the performance of face identification methods \cite{guo2019survey}
. This fact is observable when dealing with available video-surveillance datasets. While some of them, as in the case of CMU FiA \cite{goh2005fia}, DNN methods trained in general face datasets get quite good performance, even without adaptation of any kind, others, as COX Face dataset \cite{huang2015cox}, are much more challenging, and self-updating brings a considerable enhancement in performance.

\subsubsection{CMU Face in Action (FiA) database}

\label{sec:fia}

\begin{figure}[t]
    \centering
    \includegraphics[trim=30 0 10 0,width=0.32\linewidth]{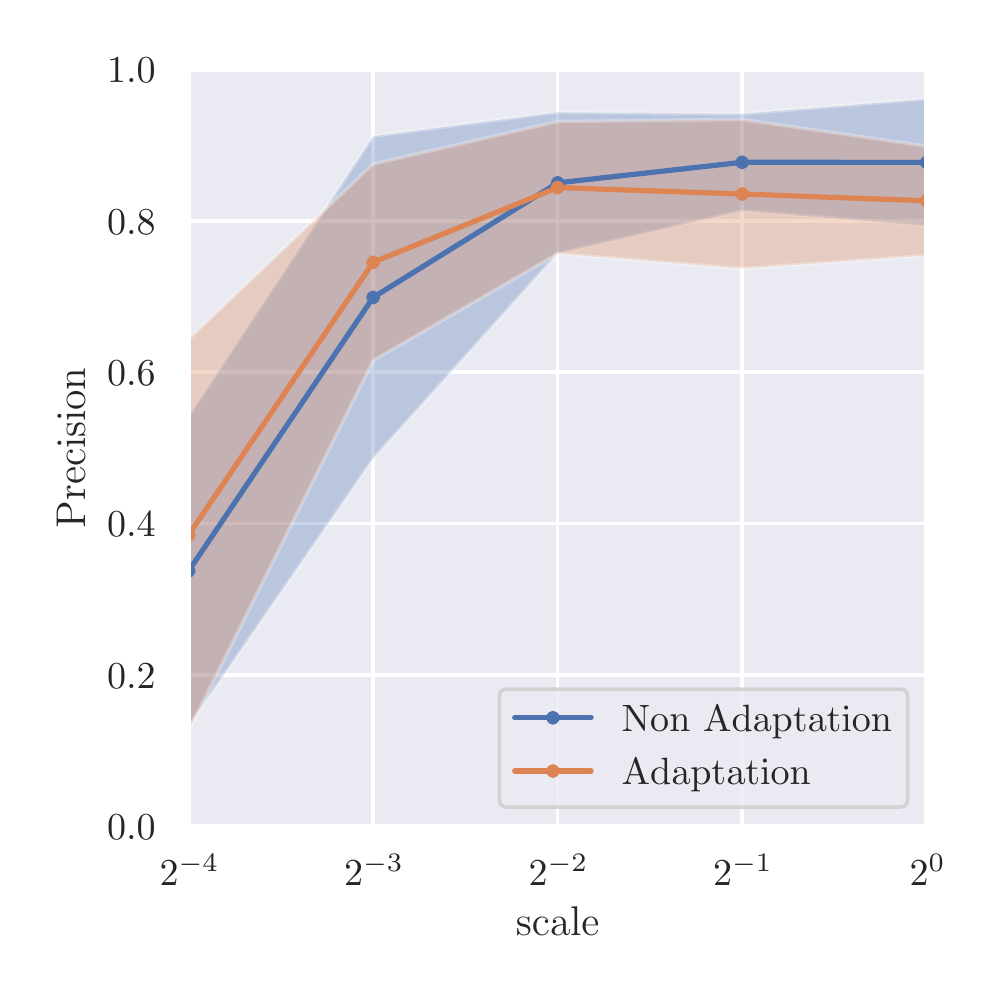}
    \includegraphics[trim=30 0 10 0,width=0.32\linewidth]{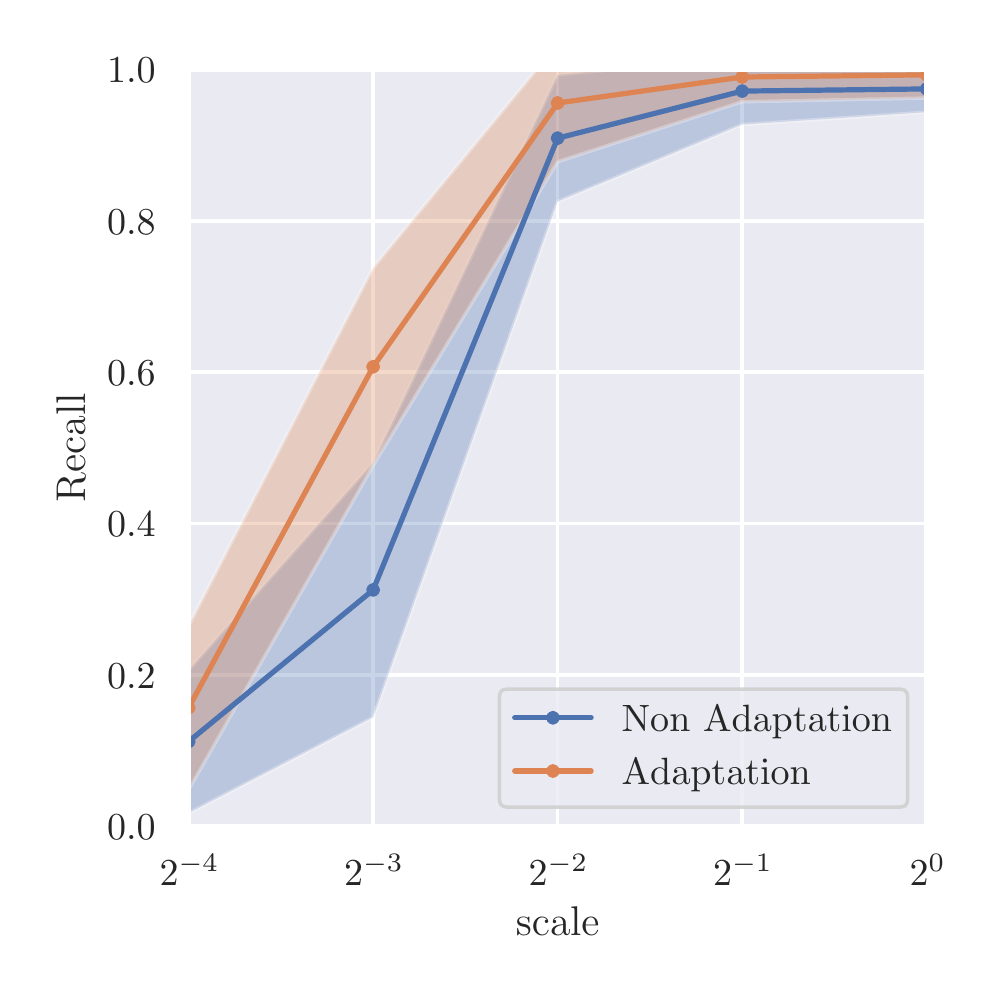}
    \includegraphics[trim=30 0 10 0,width=0.32\linewidth]{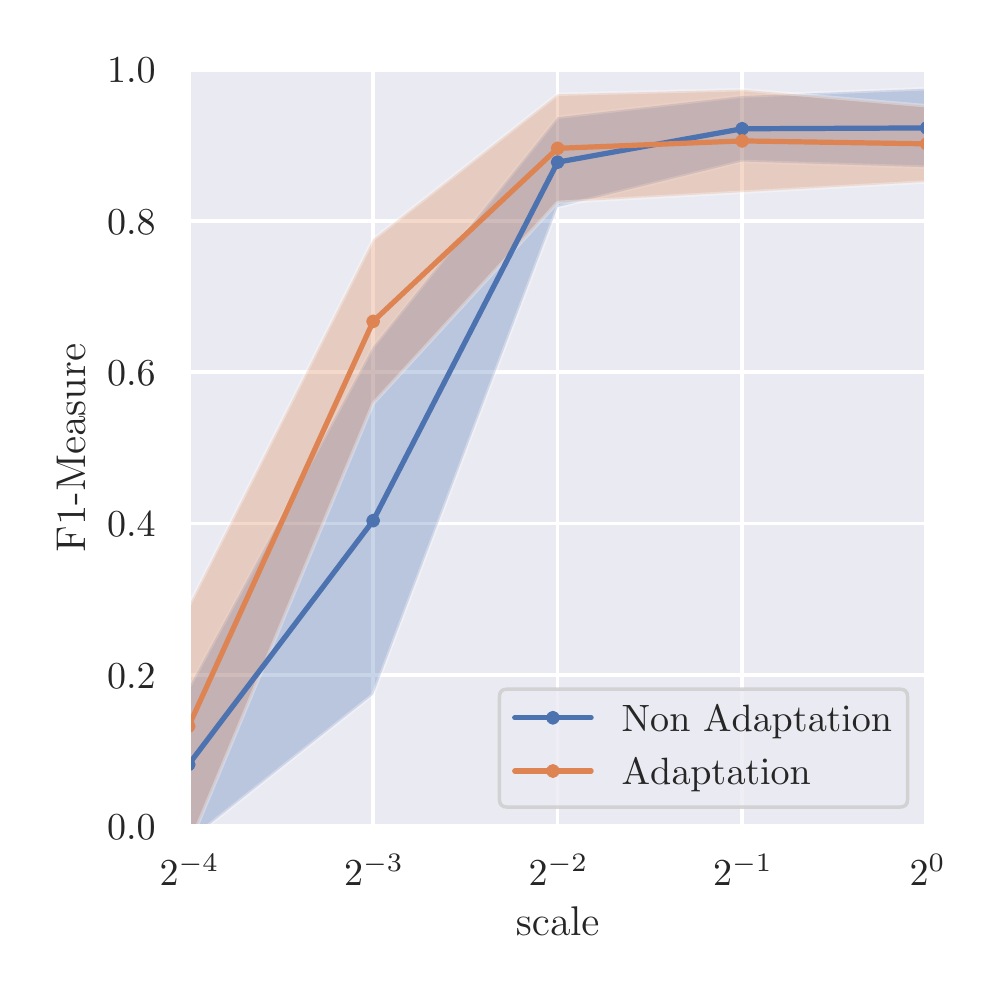}
    \caption{Performance versus image resolution, scales 1, 1/2, 1/4, 1/8 and 1/16 (that is, 112x112, 56x56, 28x28, 14x14, and 7x7 pixel image sizes). }
    \label{fig:fia}
\end{figure}

The CMU FiA database contains 20-second videos of more than 200 different individuals simulating a passport checking scenario in both indoor and outdoor environments \cite{goh2005fia}. 
Data was acquired by six synchronised cameras from 3 different angles, 2 focal lengths per angle, in 3 different sessions (3-months span between each pair of sessions). FiA video-frames present a considerable high quality, specifically in terms of resolution, since they were captured in a relatively controlled scenario. This dataset has been used to assess other adaptive methods, like the one in \cite{delatorre2015partially}.

In our experiments we have used the videos provided by the smaller focal length of the frontal camera, both indoor and outdoor,  and only considered the 70 identities present in all sessions. Given the high quality of frames, the performance of our method, even without adaptation, reached values 
of $+92\%$ in F1-score, which  widely surpasses the ones observed in \cite{delatorre2015partially}.

To challenge our method by emulating more realistic conditions, we decided to down-sample the video-frames before entering the feature encoder. In Fig.~\ref{fig:fia}  performance results for 5 different downscaling ratios are shown for the case of 35 IoI in a universe of 70 identities. Without having the possibility of averaging performance under different universes, we decided to randomly draw 20 different sets of 35 IoI for average and deviation computations. We measure OSDe-SVM performance before and after adaptation. Results on Fig.~\ref{fig:fia} show the performance degradation as the resolution decrease, which OSDe-SVM alleviates with its unsupervised adaptation. It must be taken into account that an 1/16 downscale gives face crops of size 7x7 (for an original size of 112x112); such low resolutions make identification almost unfeasible.

\subsubsection{COX Face Database}
\begin{figure}[t]
    \centering
    \includegraphics[width=\textwidth]{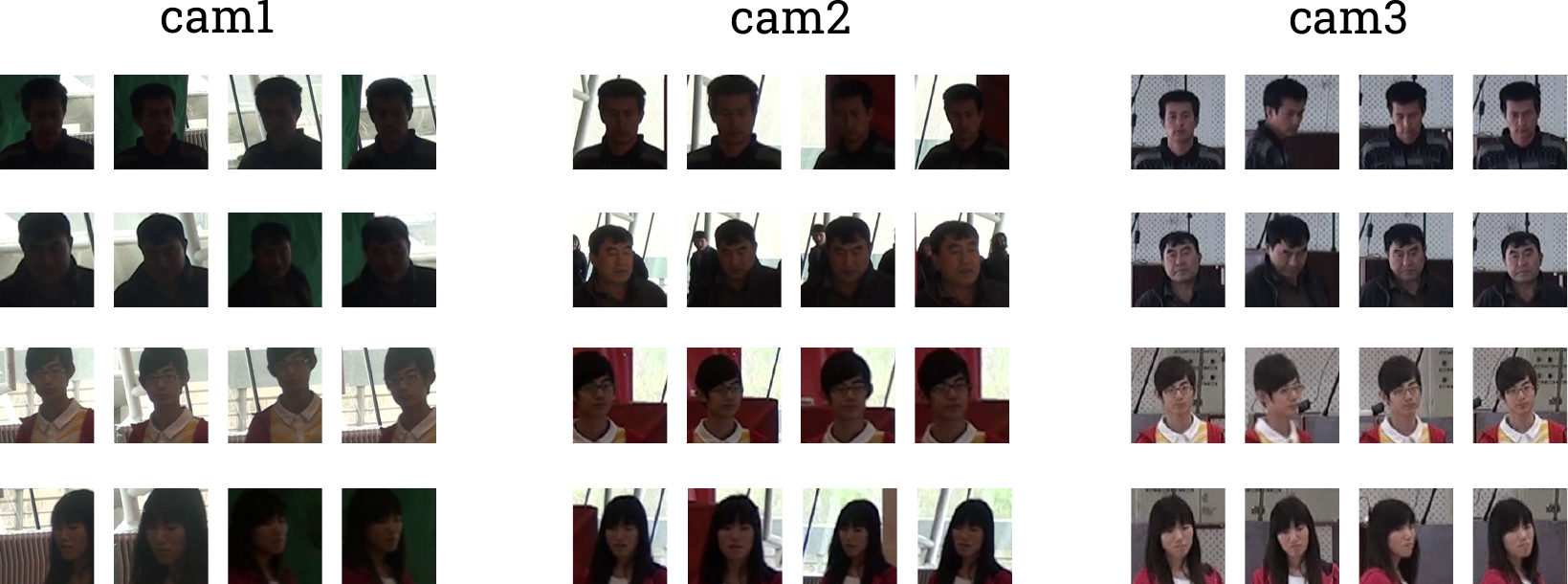}
    \caption{Samples of 4 of the 1000 identities present in the COX Face database used to perform the experiments.}
    \label{fig:examples}
\end{figure}

COX Face database \cite{huang2015cox} was specifically designed for the context of video-surveillance. There are a total of 1000 different identities in the dataset. The creators of the database asked to each of individual to follow an S-path while they capture video from 3 different viewpoints (\texttt{cam1}, \texttt{cam2} and \texttt{cam3}). Samples of each cam can be seen on Fig.~\ref{fig:examples}. Despite being taken in an interior setting, the resulting video frames present important variations in terms of both illumination and pose and especially low resolution. Samples provided by the database are the output of a commercial face tracker with a partially removed background. Nevertheless, to fine-tune this background removal and for alignment purposes, faces are passed through a face detection module for the proper performance of the feature encoder module \cite{zhang2016joint}.

Given the specific context of our application, it was necessary to perform some adjustments to adapt the data provided by COX database to how we operate. First, to increase the number of sequences per identity (and so the possibilities to update), we split each of the available videos into 3 sub-sequences keeping intact the temporal order. Second, we organised the data into different sets depending on their role in the experiments. These different sets are:

\begin{itemize}
    \item The \textbf{initially labelled sequences} are labelled video-frames of target identities used to create the first classifier of each ensemble (the sets positive samples, $P_i^k$). They consist of contains the first 5 frames from \texttt{cam1} from the 1000 individuals.
    \item The \textbf{operational sequences} simulate input sequences which would be received in the operational phase.  They consist of the three sub-sequences of \texttt{cam1} and \texttt{cam2}, and the first two sub-sequence of \texttt{cam3} of the same 1000 individuals of the \emph{initially labelled sequences}.
    \item The \textbf{testing sequences} are used to assess performance. They correspond to the last sub-sequence of \texttt{cam3} of the 1000 individuals.
\end{itemize}

Hereafter, each sequence will be noted by $S_t^k$, where $t$ refers to temporal order and $k$ refers to the identity. Following this notation, $t=0$ corresponds to the 5-frame sequences of the \emph{initially labelled sequences} used in the initialisation, $t=1,2,\dots,8$ correspond to the streaming of sequences (\emph{operational sequences}), and $t=9$ corresponds to a sequence for performance assessment (\emph{testing sequences}).

\subsection{Experimental Set-up}
\label{sec:exp-su}


We designed the experimental set-up to simulate the stream data scenario of V2V-FR. First, initial models of the IoIs are created, which consist of one-classifier ensembles. This classifier is created using samples from the \emph{initially labelled sequences} ($S_{0}^k$): 5 frames of the actual identity as a positive set and a 100 frames from other IoI as negative set (randomly drawn for each classifier). The size of the negative set is maintained for future classifier additions to have the same balance in each of the ensemble's classifiers. After this initialisation process, the system is repeatedly queried with unlabelled sequences. Since we are working in an open-set scenario, these input sequences can belong to one of the IoI or not.

\begin{figure}[t]
    \centering
    \includegraphics[width=\textwidth]{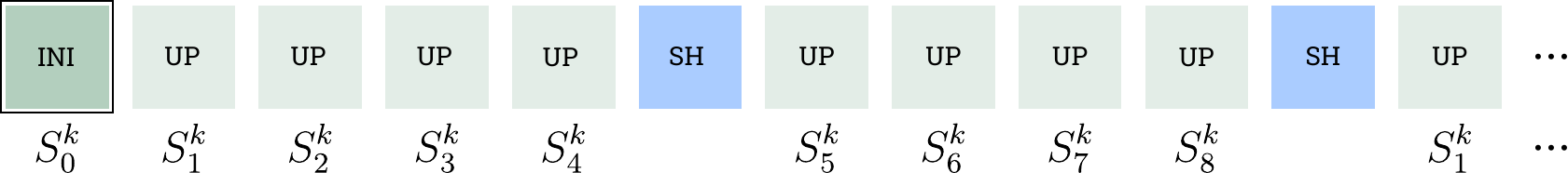}
    \caption{Adaptation steps performed during the experiments. INI stands for initialisation, UP for update and SH for self-healing. 
    The last step corresponds to the beginning of the second lap.}
    \label{fig:a_s}
\end{figure}

Experiments are organised in \emph{adaptation steps}, after which performance is measured. An \emph{adaptation step} corresponds to either the initialisation, a complete iteration over the $k$ available identities with the same $t$, or a process of self-healing (See Fig~\ref{fig:a_s}). Additionally, we fully iterate over $t=\left\{1,2,...,8\right\}$ a total of \textbf{3} times (laps), always preserving the temporal order. This way, we can increase the number of possible updates and study the system's behaviour with redundant data of both IoI and \emph{unknowns}. Self-healing was performed at \emph{adaptation steps} multiples of \textbf{5}, and the maximum number of classifiers per ensemble, $M$, was fixed to \textbf{10}. This gives us a total of \textbf{31} \emph{adaptation steps} per experiment. Alg.~\ref{algo:exp} outlines the whole procedure.

\begin{algorithm}[t]
\small
\begin{algorithmic}[1]
\State $S_{t}^k$ is the sequence $t$ of the identity $k$, $L$ is the number of laps
\State $f$ number of different sub-sequences per identity
\State  $N=$ number of IoI, $N_U=$ number identities in the universe

\For{each \emph{split}}
\For{$k = 0$ \textbf{to} $N-1$ } 
\State Initialise ensemble $k$ using $S_{0}^k$
\EndFor
\State Perform testing using the set of $S_{f}^{k=\left\{0, 1, \dots, N-1\right\}}$

\For{$lap = 0$ \textbf{to} $L-1$}
\For{$t=1$ \textbf{to} $f-1$ }


\For{$k = 0$ \textbf{to} $N_U-1$ } 
\State Perform adaptation using $S_{t}^k$.
\EndFor
\State Perform testing using the set of $S_{f}^{k=\left\{0, 1, \dots, N-1\right\}}$

\EndFor
\EndFor
\EndFor

\caption{Experimental procedure and testing protocol.}\label{algo:exp}
\end{algorithmic}
\end{algorithm}

Both the size of the identity universe and the number of IoIs vary with the experiment. For universe sizes smaller than 1000, the experiment is repeated for different splits of identities (following Alg.~\ref{algo:split}) to compute an average performance. A partial overlapping between splits was considered to get a more comprehensive sampling. For the case of 1000 identities, we repeat the experiment \textbf{5} times to address the variations provoked by the random set of negatives.  For example, for the case of a universe with 100 identities, we would have a total of 19 different splits. As for metrics, we measure precision, recall and F1-measure, using a $T_W$ fixed to 0.01.

\begin{algorithm}[t]
\small
\begin{algorithmic}[1]
  \State $N_U$ is the number of identities  in each experiment universe
  \State $N_D$ is the number of identities in the dataset
  \State i = 0
  \While{$i+N_U < N_D$}
    \State \emph{splits} $\gets$ Samples with ID $\in\left[\frac{i}{2},\frac{i}{2}+N_U\right]$
    \State$i+=N_U$
  \EndWhile
  \caption{Algorithm to create the splits}\label{algo:split}
\end{algorithmic}

\end{algorithm}

\section{Experiments and Results}
\label{sec:results}

The experimental part of the paper is organized as follows. First, we study the dependence of performance against the size of the universe, while maintaining the ratio respect to the number of IoI constant (Sec.~\ref{sec:overview}). After that, we perform a comprehensive analysis of the temporal evolution for one of the previous configurations  (Sec.~\ref{sec:temporal}).  Then, we compare the performance of our approach against state-of-the-art face recognition methods (Sec.~\ref{sec:comparison}). Finally, the effect of openness is assessed (Sec.~\ref{sec:openness}).

\subsection{Performance vs.\ Universe Size }
\label{sec:overview}

\begin{table*}
    \centering
    \footnotesize
    
    \caption{Performance over different universe sizes, while preserving the ratio with the number of IoI. 
    Values are expressed as $\mu(\sigma)$, where $\mu$ stands for mean and $\sigma$ for standard deviation.}
    \label{tab:overview}
    
    \begin{tabular}{cc ll  ll  ll }
    \toprule
          &   &  \multicolumn{2}{c}{Precision} & \multicolumn{2}{c}{Recall}  & \multicolumn{2}{c}{F1} \\
         $N$ & $N_U$ & Initial & Final & Initial & Final &Initial & Final \\
         \midrule
        
        10&20  &  75 (12)   &   71 (12)   &  79 (17)   &   88 (11)    &  76 (14)   &   78.5 (9.7)  \\
         
         20&40  & 86.9 (8.2)   &   85.1 (6.3)  & 74 (15)   &   91.1 (6.7)  &  79 (11)   &   87.8 (5.4) \\
         
         30&60  & 88.5 (6.4)   &   89.7 (5.5)  & 72 (10)    &  94.3 (3.7) &  78.8 (7.8)  &   91.8 (3.7)\\
         
         50&100 & 91.2 (4.7)   &   91.9 (3.8) &  70 (13)   &   94.2 (4.1) &  78.8 (9.0)   &   93.0 (3.2)\\
         
         100&200 &  92.6 (3.0)   &   92.6 (2.1) &  68 (10)   &   95.1 (1.9) & 77.8 (7.8)   &   93.79 (0.97)  \\
        
         200&400 & 92.0 (1.8)   &   93.5 (1.6) &  66.5 (8.5)   &   95.7 (1.0) &   76.8 (5.6)   &   94.6 (1.1)  \\
         
         300&600 &  90.3 (1.3)   &   91.9 (1.3) &   63.8 (4.8)   &   95.6 (1.0) &  74.6 (2.9)   &   93.8 (1.1) \\
         
         500&1000&  84.6 (1.4)   &   89.33 (0.76) &   63.3 (1.7)   &   95.33 (0.59) &   72.4 (1.1)   &   92.23 (0.64) \\
         
         \bottomrule
    \end{tabular}
\end{table*}

\begin{figure}[t]
    \centering
    \includegraphics[width=0.8\linewidth]{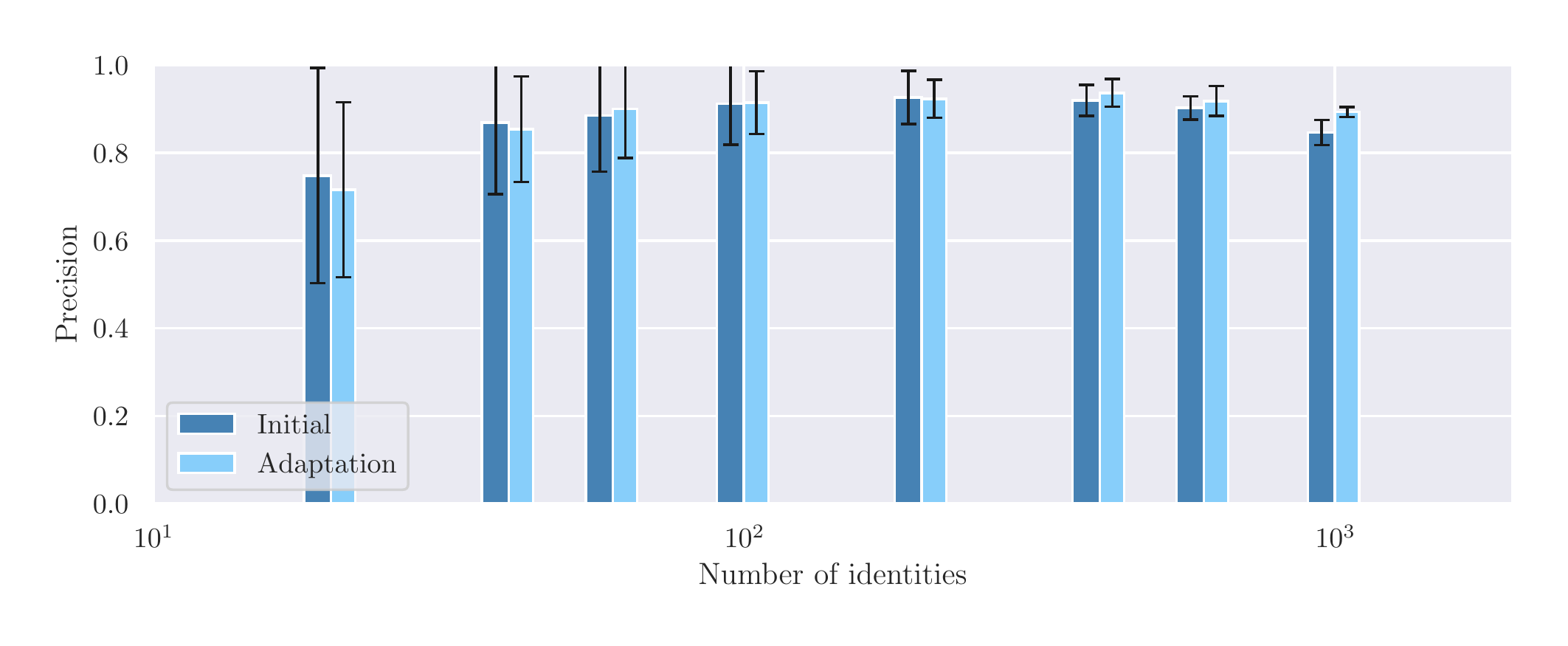}
    \includegraphics[width=0.8\linewidth]{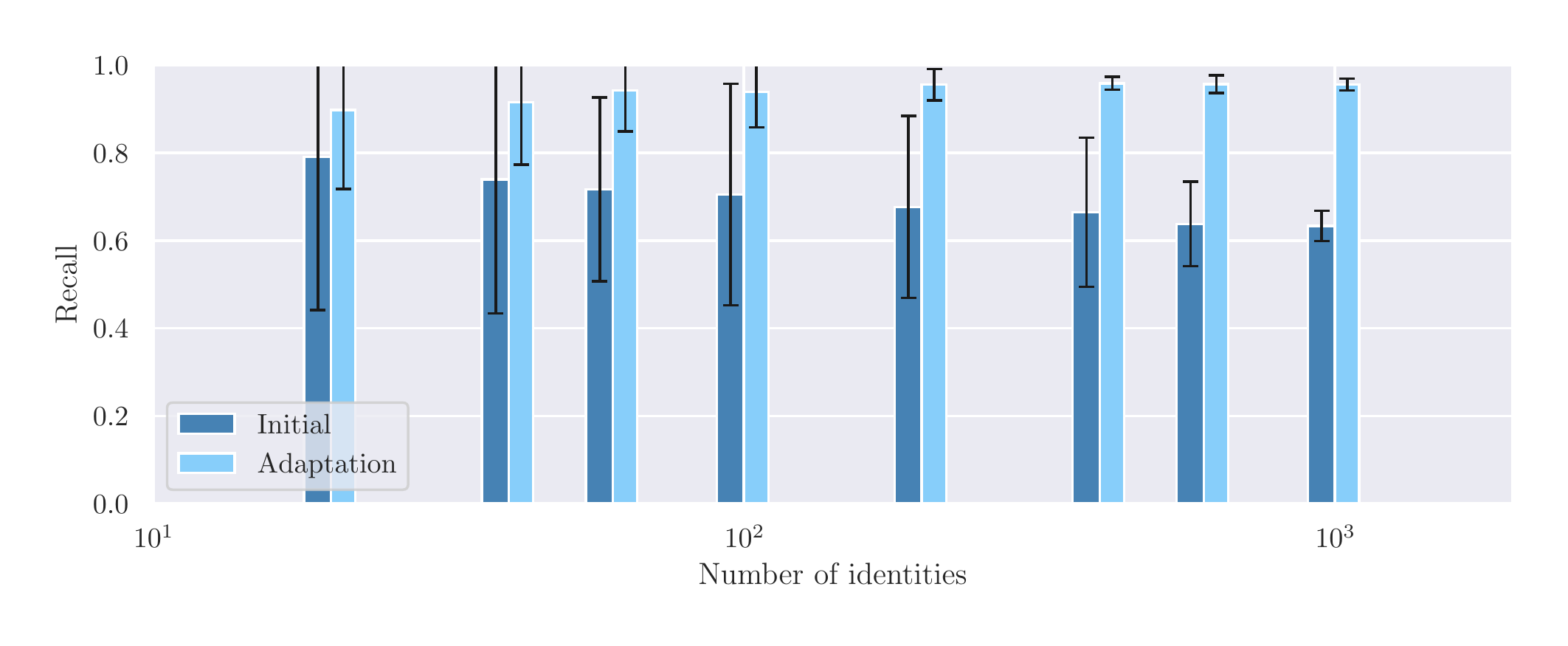}
    \includegraphics[width=0.8\linewidth]{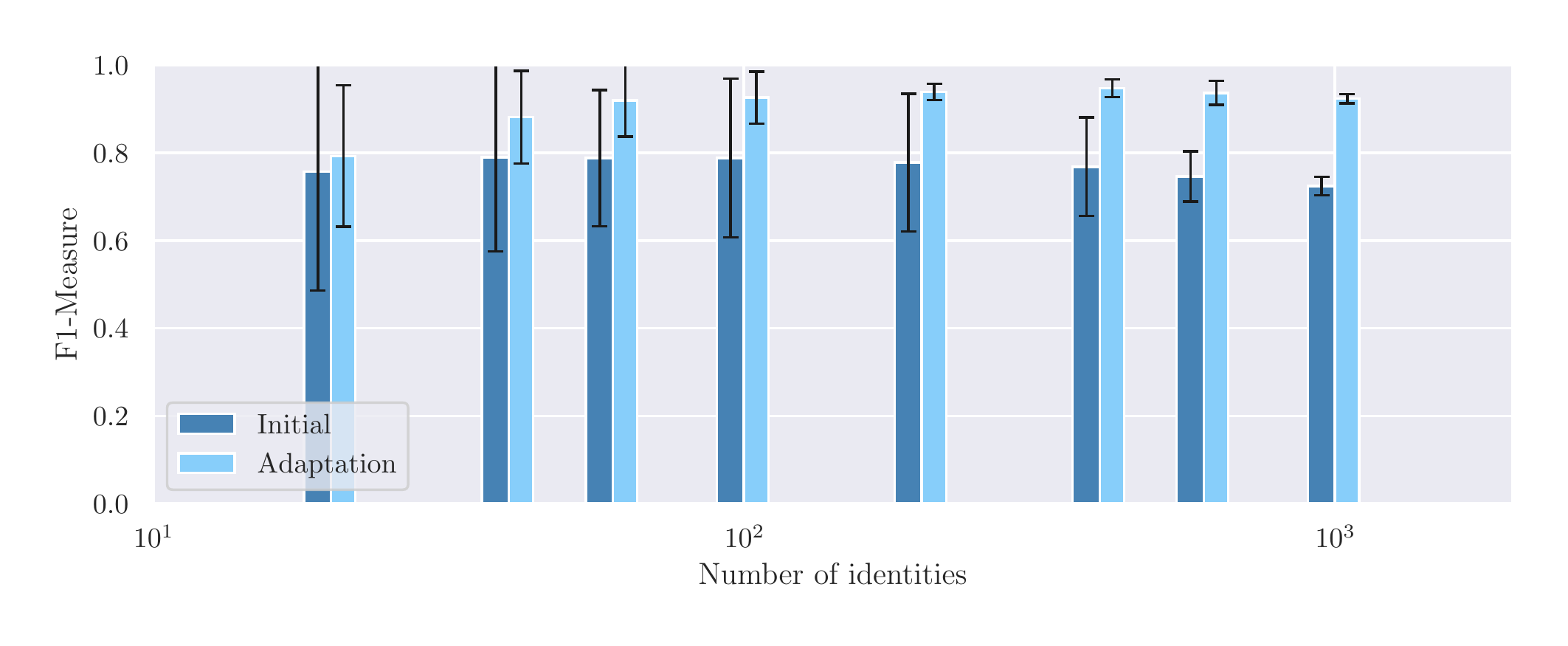}
    \caption{Performance under different universe sizes (20, 40, 60, 100, 200, 400, 600, and 1000) and same ratio  between number of IoI and universe size (1:2).}
    \label{fig:overview}
\end{figure}

This experiment shows the performance behaviour of OSDe-SVM under different universe sizes ($N_U$) while keeping the same openness index of $\approx$18\% (see Sec.~\ref{sec:openness} for further details), which corresponds to the case of having 1 IoI out of 2 identities in the universe. Results are shown in Fig.~\ref{fig:overview} and Tab.~\ref{tab:overview}. 
We measure initial (non-adaptation) and final (after adaptation) performance of OSDe-SVM, using the previously described experimental set-up (Sec.~\ref{sec:exp-su}). It is important to remark that non-adaptation means that ensembles do not incorporate new SVMs apart from the initial one. Thus, performance is quite similar to the one provided by the original network \cite{deng2019arcface}.

From the experimental results, the benefits provided by the adaptive nature of the OSDe-SVM are patent. F1-scores increase in all cases (2-20\% improvement),  mainly due to the impact on recall (9-30\% improvement). OSDe-SVM helps to enhance and enrich the existent face models, being able to recognise what previously were unrecognisable. This improvement is even more remarkable accounting the challenging experimental conditions. First, only 5 low-quality frames are provided with true labels to create the initial models. After that, no additional labelling is provided. Second, we use the same identities (both \emph{known} and \emph{unknown}) to perform the queries in each adaptation step. Therefore, confusions between identities could reinforce the impostor and eventually provoke a complete identity theft.

Although overall the behaviour observed is stable, the highest improvement in performance corresponds to larger universes. This behaviour can be explained by how we use the EVT. The quality of the Weibull fit in \emph{RDF} (Section~\ref{sec:rdf_evt}) increases as the number of samples to fit do so. For instance, since just half of the data is used in this process (those greater than the median, L8 in Alg.~\ref{algo:evt}), when the IoI is 10 the Weibull fit is done with only 5 points.

\subsection{Temporal Evolution}
\label{sec:temporal}

\begin{figure}[t]
\centering
    \subfloat[Performance   evolution.\label{sfig:hfull}]{\includegraphics[width=0.66\linewidth]{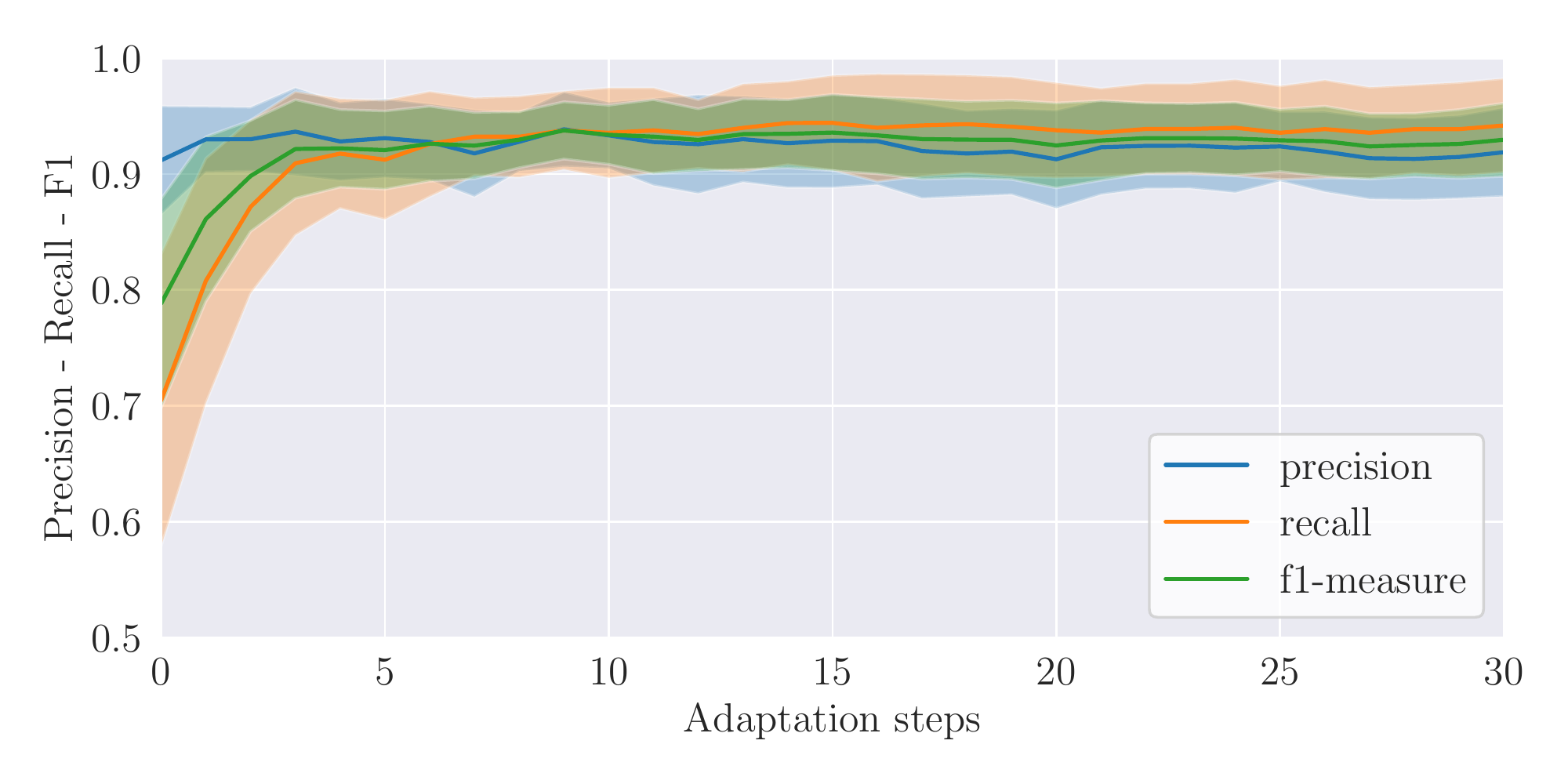}}
\hfil
    \subfloat[Ensemble size evolution.\label{sfig:hsize}]{\includegraphics[width=0.33\linewidth]{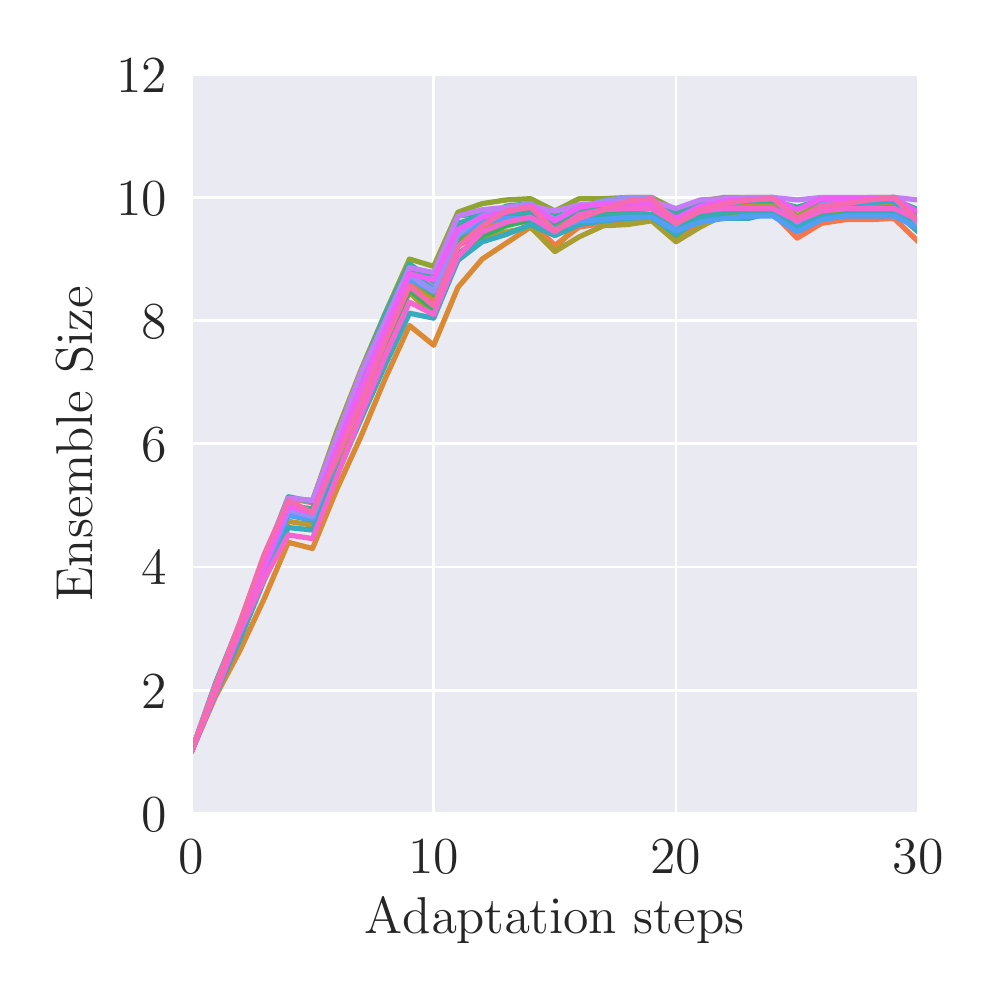}}

\caption{Evolution of OSDe-SVM for the case of 50 IoI over an universe of 100 identities.}
    \label{fig:insights}
    \end{figure}

This experiment was aimed at performing a detailed study of the temporal evolution of the OSDe-SVM performance for one of the previous cases (50 IoI in a universe of 100). Results are shown in Fig.~\ref{fig:insights}. 

The first thing we can extract from the experiments (Fig.~\ref{sfig:hfull}) is that the performance's improvement is higher in the first steps. This is something which could be expected as adding individual classifiers has a higher impact when the size of the ensemble is lower. 
Besides, this behaviour shows the system's robustness against repeated unknown queries.

These figures also allow observing in a more detailed manner the remarkable recall improvement provided by OSDe-SVM. Precision is also improved but to a lesser extent. Besides, Fig.~\ref{sfig:hsize} shows the evolution of the average ensemble size for each of the splits (Alg.~\ref{algo:split}). We can see the effect of self-healing (every 5 steps) and the limitation module. First, drops in size correspond to the triggering of the self-healing process. Second, the size of each ensemble, $M^k$, is effectively restricted by the limitation module to 10 SVM classifiers.

\subsection{Comparison against state-of-the-art face recognition models. }
\label{sec:comparison}

\begin{table}[]
    \centering
    \footnotesize
    
    \caption{Comparison against state-of-the-art face recognition models: FaceNet~\cite{schroff2015facenet} and RN100-AF (ArcFace)~\cite{deng2019arcface},  (for the case of 50 IoI in an universe of 100).} 
    \label{tab:comparison}
    
    \begin{tabular}{l rrr}
        \toprule 
        Method  & Precision & Recall & F1-measure \\
        \midrule
        FaceNet +Euclidean+TH & 38.7 (7.4) & 71.3 (9.4) & 49.0 (8.7) \\
        RN100-AF +Cosine+TH &   77.3 (9.9) & 86~ (11)~~ &  80.7 (6.5) \\
        RN100-AF +OSDe-SVM, Initial  & 91.2 (4.7) & 70~ (13)~~ & 77.8 (7.8) \\
        RN100-AF +OSDe-SVM, After Adapt. & 91.9 (3.8) & 94.2 (4.1) & 93.0 (3.2) \\
        \bottomrule
    \end{tabular}
\end{table}


Here, we compare the performance of OSDeSVM against two other well-known methods for face recognition (Tab.~\ref{tab:comparison}). In these two methods, the focus was on obtaining the most widely separated classes in feature space, to make the classification as easy as possible. On the one hand, FaceNet \cite{schroff2015facenet} feature embedding is designed to distinguish faces by computing the euclidean distance between two features (99.6\% accuracy on LFW). On the other hand, ArcFace \cite{deng2019arcface} embeddings are designed to distinguish features by using cosine similarity. All of this makes them suitable of application in any face related task (either verification, identification or general recognition) or, as our case, to use as a basis for the development of an adaptive method.

Both euclidean distance and cosine similarity are used to compare two single features. Since here we work with the features of all the frames in each query sequence, the centre of this cluster of features is computed as proposed in the original paper \cite{deng2019arcface}, to obtain a unique feature per sequence. Besides, the thresholds were tuned offline to get the best F1-scores, which are used as baselines. This would be impossible to do in stream learning conditions.

Results on Tab.~\ref{tab:comparison} allow us to gain insights into the issues addressed in this paper. First, the performance of FaceNet shows the difficult endeavour of transitioning to real-world problems (low-quality, open-set considerations, etc.). Second, our initialisation OSDe-SVM with RN100-AF embeddings preserves most of the discrimination power of the original decision function (cosine similarity). Finally, the enhanced performance provided by OSDe-SVM is put into perspective against other state-of-art static face recognition models. This improvement translates into the 15\% higher F1-score.

\subsection{Performance vs. Openness}
\label{sec:openness}

The goal of this experiment is 
to study how the behaviour of OSDe-SVM changes with the \emph{openness} ratio (\ref{eq:openness}), that is 
the ratio of \emph{known} to \emph{unknown} identities \cite{scheirer2013toward}. This measure goes from 0\% openness (closed-set recognition) to, theoretically, 100\%: 
\begin{equation}
    \textrm{openness} = 1-\sqrt{2\cdot\frac{N_\textrm{training}}{N_\textrm{target}+N_\textrm{testing}}},
    \label{eq:openness_gen}
\end{equation}
where $N_\textrm{training}$ is the number of identities used on training (in our case, $N$), $N_\textrm{target}$ is the number of identities to recognise (in our case, $N$ as well) and $N_\textrm{testing}$ are the number of identities used on testing (in our case, $N_U$). Thus, Eq.~\ref{eq:openness_gen} simplifies to:
\begin{equation}
    \textrm{openness} = 1-\sqrt{2\cdot\frac{N}{N+N_U}} .
    \label{eq:openness}
\end{equation}

\begin{figure}[t]
    \centering
    \includegraphics[trim=30 0 10 0,width=0.32\linewidth]{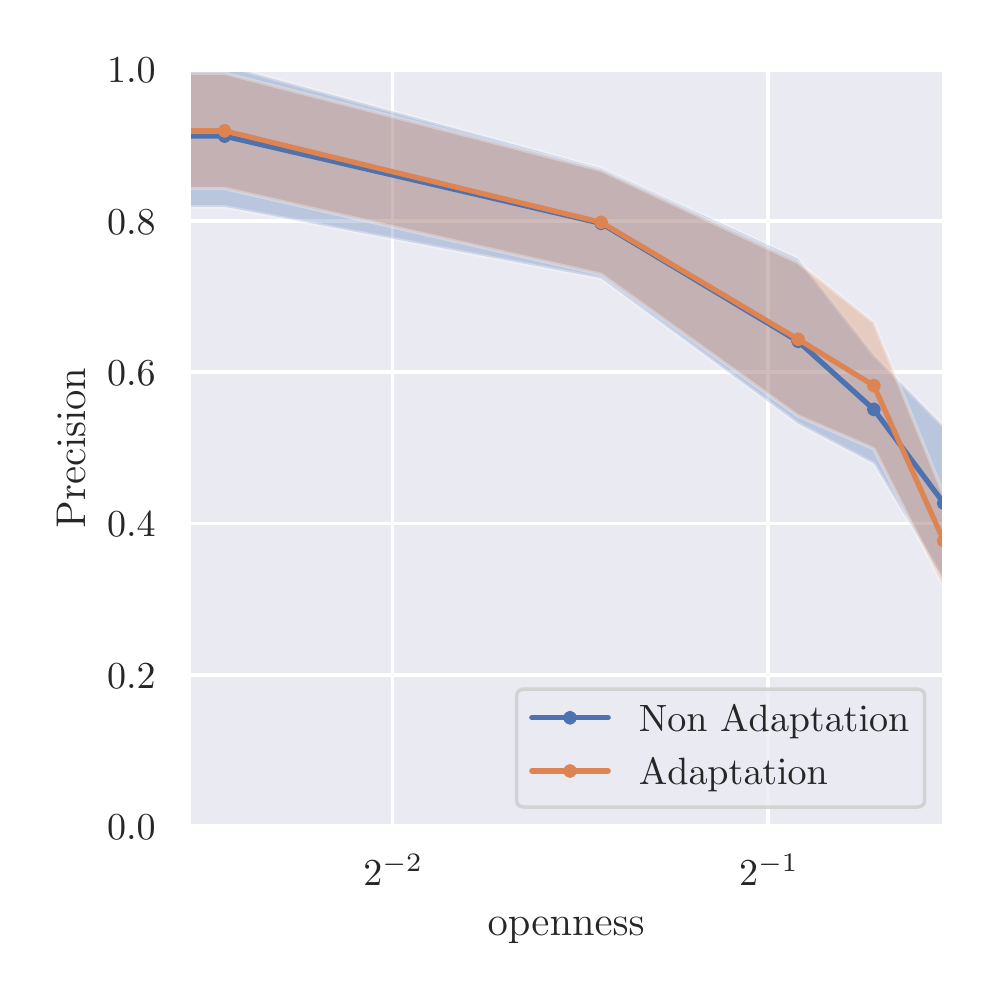}
    \includegraphics[trim=30 0 10 0,width=0.32\linewidth]{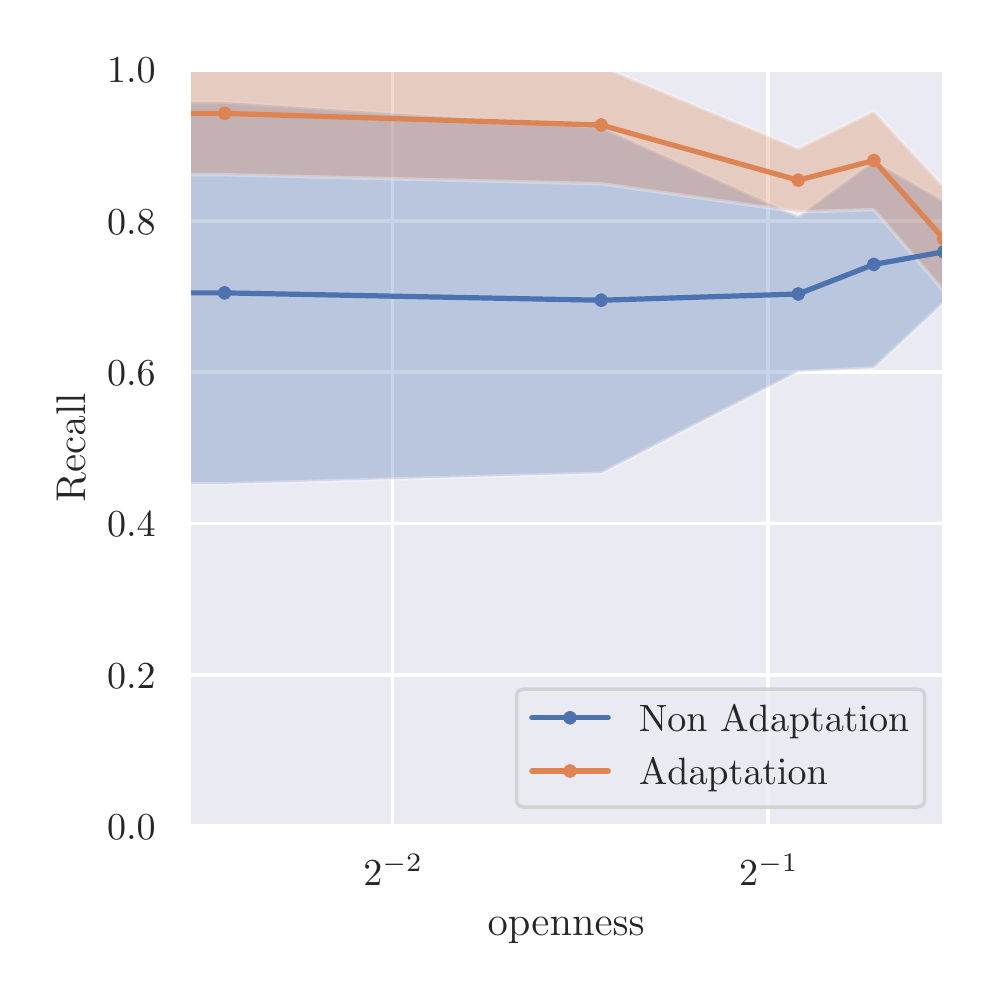}
    \includegraphics[trim=30 0 10 0,width=0.32\linewidth]{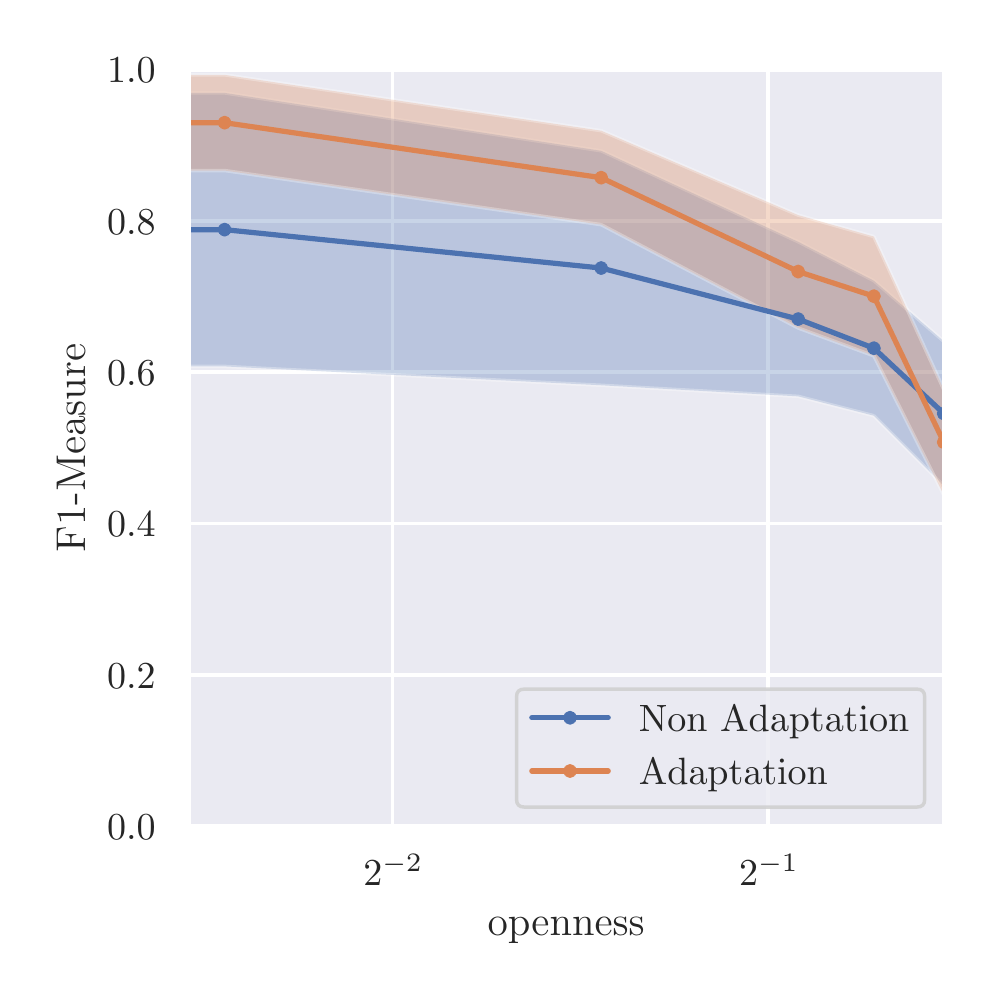}
    \caption{Performance openness dependence with fixed IoI (50), and universe $\in$ \{50, 100, 200, 400, 600, 1000\}.}
    \label{fig:openness}
\end{figure}

To have a wide range of openness values, we selected a relatively low number of IoI (50) and then vary the size of the universe from 50 identities (0\% \emph{openness}, i.e. closed-set) to 1000 identities ($\approx$70\%). Experimental results are shown in Fig.~\ref{fig:openness}, where performance is represented in terms of precision, recall and F1-scores.

The performance graphs show a clear decay of F1 performance as \emph{openness} increases,  because of the loss of precision. It  must be noted that openness affects both the unsupervised adaptation and the testing process. An increase in openness provokes a decay in precision, which also entails making more mistakes during the self-adaptation. Accordingly, the drop in precision leads to a decay in recall after the adaptation process. Against all odds, the system proves its robustness until almost a 60\% of \emph{openness}.

\section{Conclusions}
\label{sec:conclusion}

In this work, we propose a novel system, the OSDe-SVM as an instance-incremental learning approach to the problem of open-set face recognition in video surveillance. This system aims to operate in real-world non-stationary environments where availability of labelled data is quite limited. 

OSDe-SVM design uses the power of deep face representations as a basis. Once initialised (using 5 labelled frames per IoI), the proposed method creates and updates an ensemble of SVM classifiers using samples directly taken from the input sequence which effectively deals with catastrophic forgetting. These updates are performed following the self-training paradigm in which OSDe-SVM predictions are used as pseudo-labels to incorporate new knowledge without additional supervision. In this regard, we achieve update reversibility by encapsulating each update into an individual SVM classifier. By the use of EVT, OSDe-SVM can make decisions in open-set conditions.

Experiments performed on COX Face Database, to our knowledge the most challenging video-surveillance database available. Guided by real-world necessities, the set-up simulates open set recognition conditions. Results show up to a 15\% F1-measure (achieving up to a $\approx94\%$ F1-measure, depending on the amount of 
IoI to recognise) increase respect to the closest static state-of-the-art (ResNet100+AF) face recognition model. Furthermore, the proposed system's performance is tested under different degrees of \emph{openness}, proving to be reliable up to +60\% \emph{openness} (50 IoI in a universe of 1000 identities), where \emph{unknown} identities appear as many times as IoI.

As future work, apart from translating OSDe-SVM to other related machine learning applications, an interesting line of research would be to extend the proposed system to the unsupervised class-incremental problem. Following the same self-training paradigm, \emph{unknown} responses could be used to incorporate additional IoI into the recognition system. And, since classifiers are independently created, these additions would not have any adverse effect on previous knowledge.

\section*{Acknowledgements}
This work has received financial support from the Spanish government (project TIN2017-90135-R MINECO (FEDER)), from The Consellaría de Cultura, Educación e Ordenación Universitaria (accreditations 2016-2019, EDG431G/01 and ED431G/08), and reference competitive groups (2017-2020 ED431C 2017/69, and ED431C 2017/04), and from the European Regional Development Fund (ERDF). Eric López-López has received financial support from the Xunta de Galicia and the European Union (European Social Fund - ESF).









\bibliography{mybibfile}

\end{document}